\title{\textbf{Webcam-based Eye Gaze Tracking under Natural Head Movement}}
\author{Kalin M. Stefanov}

\documentclass[]{Thesis}

\usepackage{subfigure}
\usepackage{amsmath}
\usepackage{framed}
\usepackage[pdftex,colorlinks=true,citecolor=black,linkcolor=black,urlcolor=black,bookmarks=true,hypertexnames=false]{hyperref}
\usepackage{eso-pic}
\usepackage{graphicx}
\usepackage[latex]{dropping}

\newcommand{\binding}{10mm}
\renewcommand*\listfigurename{Figures}
\renewcommand*\listtablename{Tables}

\begin{document}
\pagestyle{plain}
\pagenumbering{roman}


\renewcommand{\maketitle}
{\begin{titlepage}

\let\footnotesize\small
\let\footnoterule\relax
\vspace*{-25mm}

\begin{center}  
\includegraphics{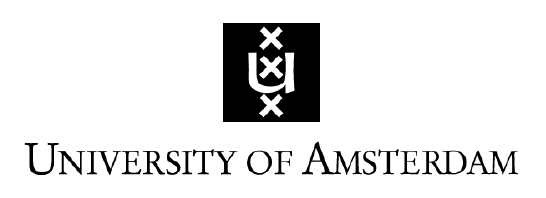} \\
\rule{100mm}{0.1mm}\medskip \\
{\Large Faculty of Science}\medskip \\
{\textsc Master in Artificial Intelligence}\medskip \\
\vskip 3.0 truecm
{\huge \bf Webcam-based Eye Gaze Tracking under Natural Head Movement \par}
\bigskip\bigskip
\vskip 1.0 truecm
\vskip 2.5 truecm
{Author: \hfill Supervisors:}
\vskip 0.3 truecm
\textbf{Kalin STEFANOV \hfill Dr. Theo GEVERS}
\vskip 0.3 truecm
\textbf{ \hfill M.Sc. Roberto VALENTI}
\vskip 1.0 truecm
\vfill
\vskip 0.7 truecm
November 2010
\vskip 0.2 truecm
\rule{40mm}{0.08mm} \\
\vskip 0.4 truecm
{Master Thesis}
\end{center}
\end{titlepage}
\cleardoublepage
}

\maketitle

\thispagestyle{empty}
\vspace*{-25mm}
\vskip 6.0 truecm
\begin{center}
\textit{``The appropriately programmed computer with the right inputs and outputs would thereby have a mind in exactly the same sense human beings have minds.''}
\vskip 1.0 truecm
John Rogers Searle
\end{center}
\cleardoublepage

\begin{abstract}

The estimation of the human's point of visual gaze is important for many applications. This information can be used in visual gaze based human-computer interaction, advertisement, human cognitive state analysis, attentive interfaces, human behavior analysis. Visual gaze direction can also provide high-level semantic cues such as who is speaking to whom, information on non-verbal communication and the mental state/attention of a human (e.g., a driver). Overall, the visual gaze direction is important to understand the human's attention, motivation and intention.

There is a tremendous amount of research concerned with the estimation of the point of visual gaze - first attempts can be traced back 1989. Many visual gaze trackers are offered to this date, although they all share the same drawbacks - either they are highly intrusive (head mounted systems, electro-ocolography methods, sclera search coil methods), either they restrict the user to keep his/her head as static as possible (PCCR methods, neural network methods, ellipse fitting methods) or they rely on expensive and sophisticated hardware and prior knowledge in order to grant the freedom of natural head movement to the user (stereo vision methods, 3D eye modeling methods). Furthermore, all proposed visual gaze trackers require an user specific calibration procedure that can be uncomfortable and erroneous, leading to a negative impact on the accuracy of the tracker. Although, some of these trackers achieve extremely high accuracy, they lack of simplicity and can not be efficiently used in everyday life.

This manuscript investigates and proposes a visual gaze tracker that tackles the problem using only an ordinary web camera and no prior knowledge in any sense (scene set-up, camera intrinsic and/or extrinsic parameters). The tracker we propose is based on the observation that our desire to grant the freedom of natural head movement to the user requires 3D modeling of the scene set-up. Although, using a single low resolution web camera bounds us in dimensions (no depth can be recovered), we propose ways to cope with this drawback and model the scene in front of the user. We tackle this three-dimensional problem by realizing that it can be viewed as series of two-dimensional special cases. Then, we propose a procedure that treats each movement of the user's head as a special two-dimensional case, hence reducing the complexity of the problem back to two dimensions. Furthermore, the proposed tracker is calibration free and discards this tedious part of all previously mentioned trackers.

Experimental results show that the proposed tracker achieves good results, given the restrictions on it. We can report that the tracker commits a mean error of (56.95, 70.82) pixels in $x$ and $y$ direction, respectively, when the user's head is as static as possible (no chin-rests are used). This corresponds to approximately $1.1^{\circ}$ in $x$ and $1.4^{\circ}$ in $y$ direction when the user's head is at distance of $750mm$ from the computer screen. Furthermore, we can report that the proposed tracker commits a mean error of (87.18, 103.86) pixels in $x$ and $y$ direction, respectively, under natural head movement, which corresponds to approximately $1.7^{\circ}$ in $x$ and $2.0^{\circ}$ in $y$ direction when the user's head is at distance of $750mm$ from the computer screen. This result coincides with the physiological bounds imposed by the structure of the human eye, where there is an inherited uncertainty window of $2.0^{\circ}$ around the optical axis.

\end{abstract}
\cleardoublepage

\tableofcontents
\addcontentsline{toc}{chapter}{Contents}

\listoffigures
\addcontentsline{toc}{chapter}{Figures}

\listoftables
\addcontentsline{toc}{chapter}{Tables}


\mainmatter
\setcounter{page}{1}

\chapter{Introduction}\label{chapter:introduction}

\dropping{2}{T}\newline he modern approach to artificial intelligence is centered around the concept of a \textit{rational agent}. An agent is anything that can \textit{perceive} its environment through sensors and \textit{act} upon that environment through actuators \cite{RUSSELL&NORVIG}. An agent that always tries to optimize an appropriate performance measure is called rational. Such a definition of a rational agent is fairly general and it can include human agents (having eyes as sensors and hands as actuators), robotic agents (having cameras as sensors and wheels as actuators), or software agents (having a graphical user interface as sensors and as actuators). From this perspective, artificial intelligence can be regarded as the study of the principles and design of artificial rational agents.

The interaction between the human and the world occurs through information being received and sent: \textit{input} and \textit{output}. When a human interacts with a computer he/she receives output information from the computer, and responds to it by providing input to the computer - the human's output becomes the computer's input and vice versa. For example, sight is used by the human primarily in receiving information from the computer, but it can also be used to provide information to the computer.

Input in the human occurs mainly through the senses and output through the motor control of the effectors. There are five major senses: sight, hearing, touch, taste and smell. The first three of these are the most important to human-computer interaction. Taste and smell do not currently play a significant role in human-computer interaction, and it is not clear whether they could be exploited at all in general computer systems. However, vision, hearing and touch are central. Similarly, there are number of effectors: limbs, fingers, eyes, head and vocal system. The fingers play the primary role in the interaction with the computer, through typing and/or mouse control.

Imagine using a personal computer with a mouse and a keyboard. The application you are using has a graphical user interface, with menus, icons and windows. In your interaction with the computer you receive information primary by sight, from what appears on the screen. However, you may also receive information by hearing: for example, the computer may `beep' at you if you make a mistake. Touch plays a part in that you feel the keys moving (also you hear the `clicks'). You yourself, are sending information to the computer using your hands, either by hitting the keys or moving the mouse. Sight and hearing do not play a direct role in sending information to the computer in this example, although they may be used to receive information from a third source (for example, a book, or the words of another person), which is then transmitted to the computer.

Since the introduction of the computers, advances in communication with the human, have mainly been made on the communication from the computer to the human (graphical representation of the data, window systems, use of sound), whereas communication from the human to the computer is still restricted to keyboards, joysticks and mice - all operated by hand. By tracking the direction of the visual gaze of the human for example, the bandwidth of communication from the human to the computer can be increased by using the information about what the human is looking at. This is only an example for possible direction towards increasing the human-computer bandwidth; by monitoring the entire state of the human, the computer can react to all kinds of gestures and/or voice commands.

This leads to a new way of regarding the computer, not as a tool that must be operated explicitly by commands, but instead as an \textbf{agent} that \textbf{perceives} the human and \textbf{acts} upon that perception. In turn, the human is granted with the freedom to concentrate on interacting with the data presented by the computer, instead of using the computer applications as tools to operate on the data.

As stated in \cite{NISHINO&NAYAR}, ``Our eyes are crucial to us as they provide us an enormous amount of information about our physical world'', i.e., vision is an essential tool in interaction with other human beings and the physical world that surrounds us. If computers were able to understand more about our attention as we move in the world, they could enable simpler, more realistic and intuitive interaction. To achieve this goal, simple and efficient methods of analyzing human cues (body language, voice tone, eye fixation) are required, that would enable computers to interact with humans in more intelligent manner.

We envision that a future computer user who would like to, say, use some information from the Internet for doing some work, could walk into the room where the \textit{computer agent} is located and interact with it by looking, speaking and gesturing. When the user's visual gaze falls on the screen of the computer agent, it would start to operate from the user's favorite starting point or where he/she left off the last time.

Wherever the user looks, the computer agent will begin to emphasize the appropriate data carrying object (database information, ongoing movies, videophone calls), and utterances like `more' combined with a glance or a pointing gesture will zoom in on the selected object. If the user's visual gaze flutters over several things, the computer agent could assume that the user might like an overview, and an appropriate zooming out or verbalized data summery can take place.

To investigate the present status of technology needed for this class of futuristic, multi-modal systems, we have chosen to focus on one of the interaction techniques discussed, namely the visual gaze interaction. It is evident that the estimation of the human's point of visual gaze is important for many applications. We can apply such system in visual gaze based human-computer interaction, advertisement \cite{SMITH&BA&ODOBEZ&GATICA-PEREZ}, human cognitive state analysis, attentive interfaces (visual gaze controlled mouse), human behavior analysis. Visual gaze direction can also provide high-level semantic cues such as who is speaking to whom, information on non-verbal communication (interest, pointing with the head, pointing with the eyes) and the mental state/attention of a human (e.g., a driver). Overall, the visual gaze direction is important to understand the human's attention, motivation and intention \cite{HANSEN&JI}.

\section{Problem Definition}\label{sec:problem}

A multi-modal human-computer interface is helpful for enhancing the human-computer communication by processing and combining multiple communication modalities known to be helpful in human-human interaction. Many human-computer interaction applications require the information where the human is looking at, and what he/she is paying attention to. Such information can be obtained by tracking the orientation of the human's head and visual gaze. While current approaches to visual gaze tracking tend to be highly intrusive (the human must either be perfectly still, or wear a special device), in this study we present a non-intrusive appearance-based visual gaze tracking system.

We are faced with the following scenario - the human sits in front of the computer and looks around on the screen. We would like to be able to determine the point on the screen at which the human is looking without imposing any restrictions. These could be - special hardware (light sources, expensive cameras, chin-rests), or any prior knowledge for the scene set-up (human's position with respect to the screen and the camera), or limiting the human to be as static as possible (no head movements). Let's point out some usability requirements for our visual gaze tracking system. According to \cite{MORIMOTO&MIMICA} an ideal visual gaze tracker should:

\begin{itemize}
\item{be accurate, i.e., precise to minutes of arc;}
\item{be reliable, i.e., have constant, repetitive behavior;}
\item{be robust, i.e, should work under different conditions, such as indoors and outdoors, for people with glasses and contact lenses;}
\item{be non-intrusive, i.e., cause no harm or discomfort;}
\item{allow for free head motion;}
\item{not require calibration, i.e., instant set-up;}
\item{have real-time response.}
\end{itemize}

Therefore, it would be interesting to develop a visual gaze tracking system using only a computer and a digital video camera, e.g., a web camera. In such system the input in the human occurs through vision (what is displayed on the computer screen - computer's output) and the output in the human occurs through the point of visual gaze - computer's input through the web camera. This would lead to truly non-intrusive visual gaze tracking and a more accessible and user friendly system.

The human's visual gaze direction is determined by two factors: the position of the head, and the location of the eyes \cite{LANGTON&HONEYMAN&TESSLER}. While the position of the head determines the overall direction of the visual gaze, the location of the eyes determines the exact visual gaze direction and is limited by the head position (human's field of view; see section \ref{sec:experimentsConclusions}). In this manuscript, we propose a system that first estimates the position of the head to establish the overall direction of the visual gaze and then it estimates the location of the eyes to determine the exact visual gaze direction restricted by the current head position.

\begin{figure}[htp!]
\begin{center}
\includegraphics[scale=1]{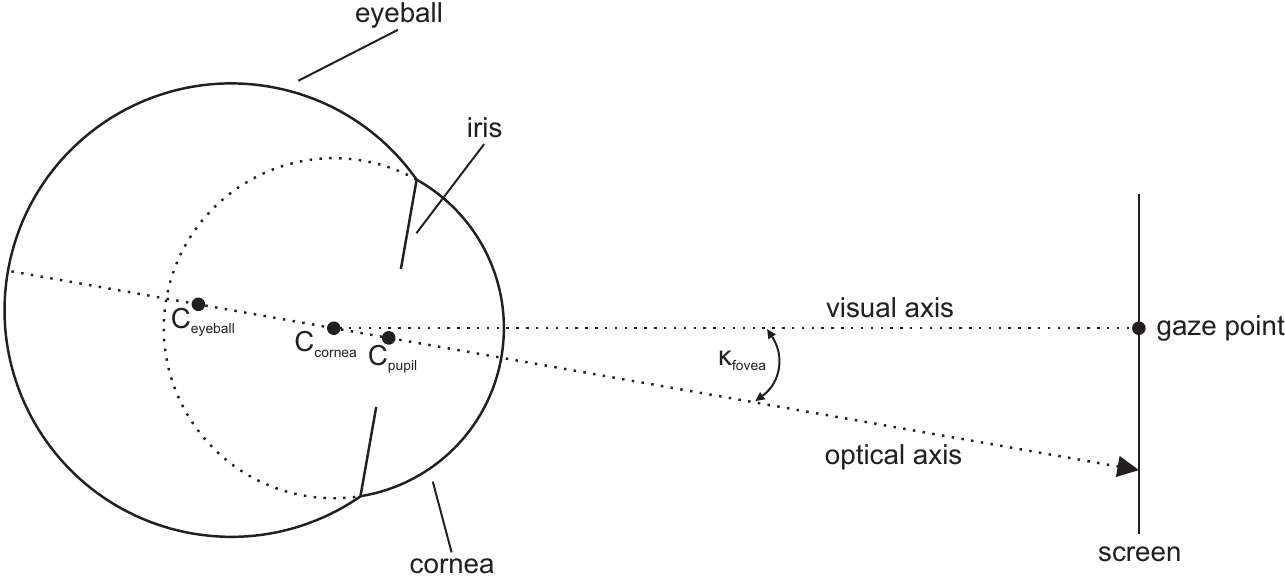}
\caption{Definition of the point of visual gaze}
\label{fig:gazeDefinitionIntro}
\end{center}
\end{figure}

We start our investigation for possible solutions by defining the point of visual gaze. In 3D space the point of visual gaze is defined as the intersection of the visual axis of the eye with the computer screen (see figure \ref{fig:gazeDefinitionIntro}), where the optical axis of the eye is defined as the 3D line connecting the corneal center, $C_{cornea}$, and the the pupil center, $C_{pupil}$. The angle between the optical axis and the visual axis is named \textit{kappa} ($\kappa_{fovea}$) and it is a constant value for each person.

\section{Previous Work}\label{sec:background}

\begin{figure}[htp!]
\begin{center}
\includegraphics[scale=1]{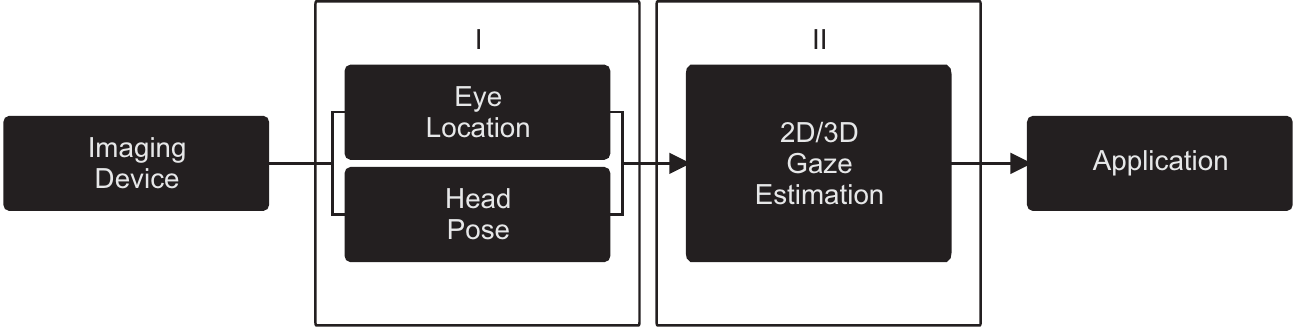}
\caption{The visual gaze estimation pipeline}
\label{fig:gazePipeline}
\end{center}
\end{figure}

Usually, the procedure of estimating the point of visual gaze consists of two steps (see figure \ref{fig:gazePipeline}): in the first step (I) we analyze and transform pixel based image features obtained by a device (camera) to a higher level representation (the position of the head and/or the location of the eyes) and then (II), we map these features to estimate the visual gaze direction, hence finding the area of interest on the computer screen.

The literature is rich with research concerned with the first component of the visual gaze estimation pipeline, that is covered by methods for estimation of the head position and the eyes location. Good detection accuracy is achieved by using intrusive or expensive sensors which often limit the possible movement of the user's head, or require the user to wear a device \cite{BATES&ISTANCE&OOSTHUIZEN&MAJARANTA}. Eye center locators that are solely based on appearance are proposed by \cite{CRISTINACCE&COOTES&SCOTT}, \cite{KROON&BOUGHORBEL&HANJALIC}, \cite{VALENTI&GEVERS}, and they are reaching reasonable accuracy in order to roughly estimate the area of attention on the computer screen in the second component of the visual gaze estimation pipeline. A recent survey \cite{HANSEN&JI} discusses the different methodologies to obtain the eyes location information through video-based devices. The survey in \cite{MURPHY-CHUTORIAN&TRIVEDI} gives a good overview of appearance based head pose estimation methods.

Once the correct features are determined using one of these methods, the second component in the visual gaze estimation pipeline maps the obtained information to the 3D scene in front of the user. In visual gaze trackers, this is often achieved by direct mapping of the location of the center of the eyes to a location on the computer screen. This requires the system to be calibrated and often limits the possible position of the user's head (e.g., by using chin-rests). In case of 3D visual gaze estimation, this often requires the intrinsic camera parameters to be known. Failure to correctly calibrate or comply with the restrictions of the visual gaze tracker may result in wrong estimation of the point of visual gaze. 

Recently, a third step in the visual gaze estimation pipeline was proposed by \cite{VALENTI&VALENTI}. The author describes how the saliency framework can be exploited in order to correct the estimate for the point of visual gaze; the current picture on the computer screen is used as a probability map and the estimated point of visual gaze is steered towards the closest \textit{interesting} point. Furthermore, a procedure for correcting the errors introduced by the calibration step is described.

\begin{figure}[htp!]
\begin{center}
\includegraphics[scale=1]{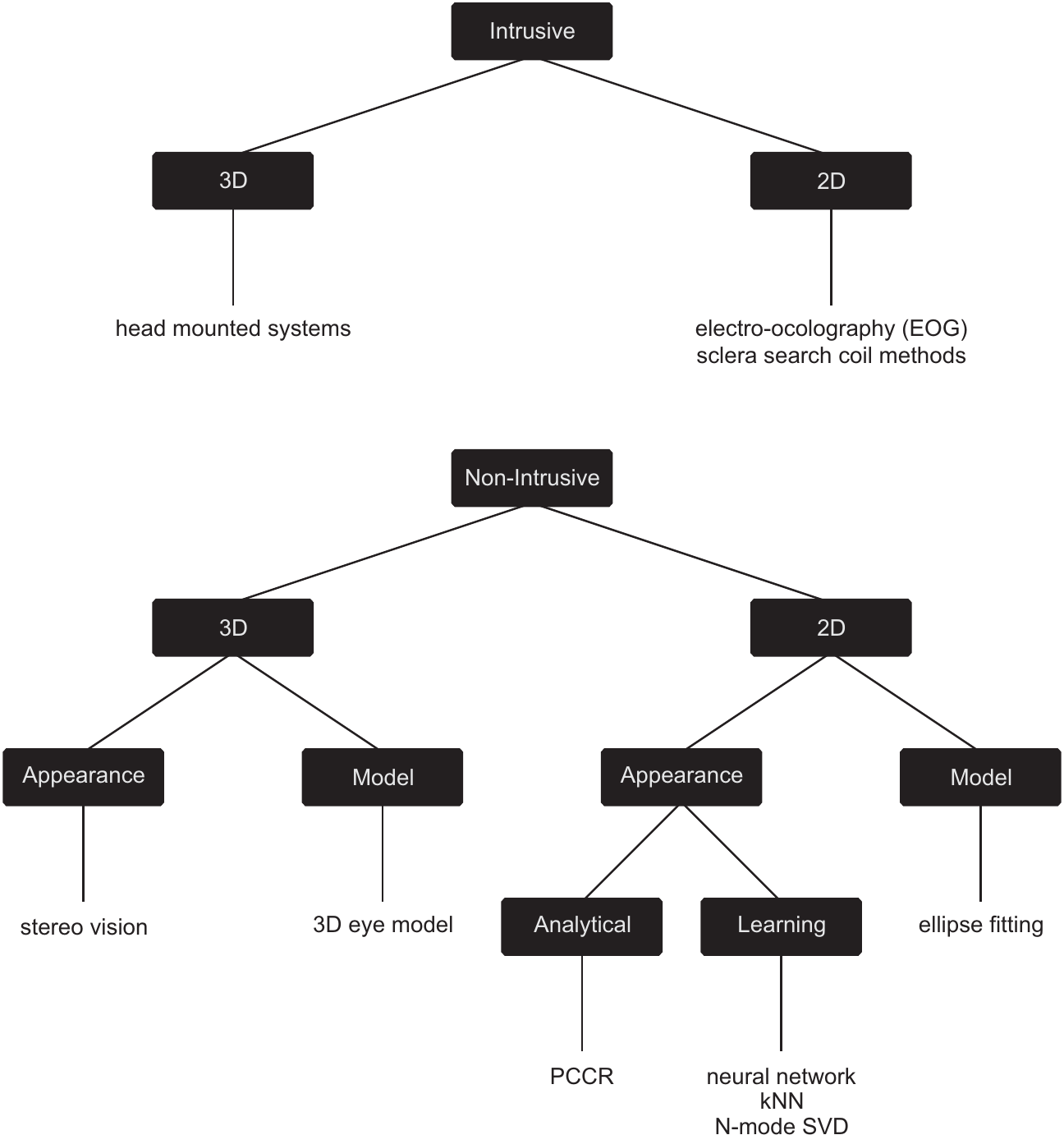}
\caption{Classification of different eye tracking techniques}
\label{fig:eyeTrackingTechniques}
\end{center}
\end{figure}

Since the accurate estimation of the point of visual gaze is controlled by the location of the eyes, the approach one takes for estimation of the location of the eyes, is the one that will define the type of the visual gaze tracker. Figure \ref{fig:eyeTrackingTechniques} offers a classification of different eye tracking techniques; these can be divided into two big classes - intrusive and non-intrusive, where each class is further subdivided.

\newpage

We refer the reader to \cite{ROBINSON}, \cite{KAUFMAN&BANDOPADHAY&SHAVIV} and \cite{ESTRANY&FUSTER&GARCIA&LUO} for examples of intrusive approaches for visual gaze estimation. \cite{HEINZMANN&ZELINSKY}, \cite{CHEN&JI}, \cite{SHIH&WU&LIU}, \cite{YOSHIO&ALEXANDER}, \cite{SHIH&LIU}, \cite{KOHLBECHER&BARDINST&BARTL&SCHNEIDER&POITSCHKE&ABLASSMEIER}, \cite{BEYMER&FLICKNER}, \cite{PARK&LEE&KIM} and \cite{NEWMAN&MATSUMOTO&ROUGEAUX&ZELINSKY} offer different 3D solutions of the problem. The most popular techniques found in the literature are based on the PCCR (Pupil Center-Corneal Reflection) technique. Examples of such systems can be found in \cite{OHNO&MUKAWA}, \cite{YOO&CHUNG1}, \cite{PEREZ&CORDOBA&GARCIA&MENDEZ&MUNOZ&PEDRAZA&SANCHEZ}, \cite{HENNESSEY&NOUREDDIN&LAWRENCE}, \cite{ZHU&JI1}, \cite{ZHU&JI2}, \cite{MORIMOTO&KOONS&AMIR&FLICKNER}, \cite{OHNO&MUKAWA&YOSHIKAWA}, \cite{YOO&LEE&CHUNG}, \cite{YOO&CHUNG2}, \cite{ZHU&JI3}, \cite{RYOUNG&JAIHIE} and \cite{LIU&JIN&WU}. There are examples of learning algorithms, \cite{PIRATLA&JAYASUMANA}, \cite{SUGANO&MATSUSHITA&SATO&KOIKE}, \cite{NORIS&BENMACHICHE&BILLARD}, \cite{PARK&LEE&KIM&LECLAIR} and \cite{STIEFELHAGEN&YANG&WAIBEL}, although they reach poorer accuracy than the previously mentioned approaches.

\section{Our Method}

Figure \ref{fig:eyeTrackingTechniques} gives the general overview of the eye tracking techniques in the literature. In order to comply with the usability requirements outlined earlier (see section \ref{sec:problem}) we propose a visual eye-gaze tracker that can be classified as a non-intrusive, 3D, and appearance based. The proposed tracker treats the movement of the user's head as a special two-dimensional case, and as such, it is able to reduce the three-dimensional problem of visual eye-gaze estimation to two dimensions and exploit the simplicity of the solution there.

\section{Overview}\label{sec:overview}

The rest of this manuscript is structured as follows:

\begin{itemize}
\item{chapters \ref{chapter:SH} and \ref{chapter:NHM} offer deeper investigation of the problem. The chapters give detailed description of the proposed visual eye-gaze trackers;}
\item{chapter \ref{chapter:experiments} describes the experiments conducted to test our ideas. The chapter offers observations section which outlines better experiment methodology and argues that the experiments conducted bear the power of real life system usage;}
\item{chapter \ref{chapter:results} presents the results of the performed experiments. It offers a discussion on the results and the origins of errors;}
\item{chapter \ref{chapter:conclusions} concludes and summarizes the proposed approach. Possible improvements, future work and main contributions are outlined.}
\end{itemize}
\cleardoublepage

\chapter{No Head Movement}\label{chapter:SH}

\dropping{2}{I}\newline n this chapter we propose a visual eye-gaze tracker that is capable to estimate and track the user's point of visual gaze on the computer screen, using a single low resolution web camera, under the constraint that the user's head is static. The tracker proves that accurate tracking of the user's point of visual gaze in this settings is possible, through simple and efficient computer vision algorithms. The tracker implements a procedure that is suitable for situations when the user's head is as static as possible and it is regarded as \textit{2D} in the rest of the manuscript.

This tracker can be classified as a non-intrusive, 2D, and appearance based. However, it is not based on the PCCR (Pupil Center-Corneal Reflection) technique, but solely on image analysis. Generally, the tracker is composed of four steps (two steps in each of the components of the visual gaze estimation pipeline shown in figure \ref{fig:gazePipeline}). The first component of the pipeline combines an algorithm that detects and tracks the location of the user's eyes, and an algorithm that defines and tracks the location of user's facial features; the facial features play the role of anchor points and eyes displacement vectors that are used in the second component of the pipeline are constructed between the location of the facial features and the location of the eyes. The second component of the pipeline combines a calibration procedure that gives correspondences between facial feature-pupil center vectors and points on the computer screen (train data), and a method (see appendix \ref{chapter:AppendixB}) to approximate a solution of overdetermined system of equations for future visual gaze estimation.

\section{Tracking the Eyes}\label{sec:eyes}

Since we assume that the user's head is static in this scenario, it is evident that, the accurate detection and tracking of the location of the eyes is at most importance for the performance of the 2D visual eye-gaze tracker. Our first attempt for eye location estimation and tracking is based on the template matching technique - finding small parts of an image which match a template image.

This method is normally implemented by first picking out a part of the search image to use as a template. Template matching matches an actual image patch against an input image by `sliding' the patch over the input image using a matching method. For example, if, we have an image patch containing a face then, we can slide that face over an input image looking for strong matches that would indicate that another face is present. Let's use $I$ to donate the input image (where we are searching), $T$ the template, and $R$ the result. Then,

\begin{itemize}
\item{\textbf{squared difference} matching method matches the squared difference, so a perfect match will be 0 and bad match will be large:

\begin{equation}
R_{sq\_diff}(x,y) = \displaystyle\sum_{x',y'}[T(x',y') - I(x + x',y + y')]^{2}
\end{equation}}
\item{\textbf{correlation} matching method multiplicatively matches the template against the image, so a perfect match will be large and bad match will be small or 0:

\begin{equation}
R_{ccorr}(x,y) = \displaystyle\sum_{x',y'}[T(x',y')\cdot I(x + x',y + y')]^{2}
\end{equation}}
\item{\textbf{correlation coefficient} matching method matches a template relative to its mean against the image relative to its mean, so a perfect match will be 1 and a perfect mismatch will be -1; a value of 0 simply means that there is no correlation (random alignments):

\begin{equation}
R_{ccoeff}(x,y) = \displaystyle\sum_{x',y'}[T'(x',y')\cdot I'(x + x',y + y')]^{2}
\end{equation}

\begin{equation}
T'(x',y') = T(x',y') - \frac{1}{(w\cdot h)\displaystyle\sum_{x'',y''}T(x'',y'')}
\end{equation}

\begin{equation}
I'(x + x',y + y') = I(x + x',y + y') - \frac{1}{(w\cdot h)\displaystyle\sum_{x'',y''}I(x + x'',y + y'')}
\end{equation}}
\item{\textbf{normalized methods} - for each of the three methods just described, there are also normalized versions first developed by Galton \cite{GALTON} as described by Rodgers \cite{RODGERS&NICEWANDER}. The normalized methods are useful because, they can help to reduce the effects of illumination differences between the template and the search image. In each case, the normalization coefficient is the same:

\begin{equation}
Z(x,y) = \sqrt{\displaystyle\sum_{x',y'}T(x',y')^{2}\cdot \displaystyle\sum_{x',y'}I(x + x',y + y')^{2}}
\end{equation}

The values for each method that give the normalized computations are:

\begin{equation}
R_{sq\_diff\_normed}(x,y) = \frac{R_{sq\_diff}(x,y)}{Z(x,y)}
\end{equation}

\begin{equation}
R_{ccor\_normed}(x,y) = \frac{R_{ccor}(x,y)}{Z(x,y)}
\end{equation}

\begin{equation}
R_{ccoeff\_normed}(x,y) = \frac{R_{ccoeff}(x,y)}{Z(x,y)}
\end{equation}}
\end{itemize}

The template matching technique has the advantage to be intuitive and easy to implement. Unfortunately, there are important drawbacks associated with it: first, the template needs to be selected by hand every time the tracker is used, and second, the accuracy of the technique is generally low. Furthermore, the accuracy is dependent on the illumination conditions: too different illumination conditions between the template and the search image lead to a negative impact on the accuracy of estimate for the location of the eye.

\begin{figure}[htp!]
\begin{center}
\subfigure[]{
\includegraphics[scale=2]{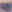}}
\subfigure[]{
\includegraphics[scale=2]{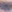}}
\subfigure[]{
\includegraphics[scale=0.5]{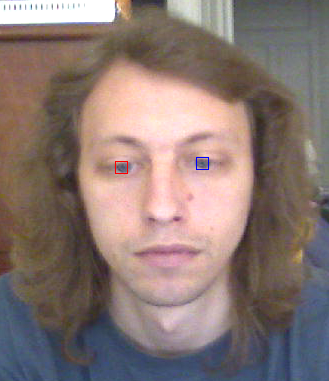}}
\caption{Template matching based on the squared difference matching method}
\label{fig:templateMatching}
\end{center}
\end{figure}

Figure \ref{fig:templateMatching} illustrates the procedure of template matching based on the squared difference matching method: left eye template (a), right eye template (b), and their matches in the search image (c). The eye regions are selected at the beginning of the procedure and then, the regions with the highest response to the templates are found in the search image.
\newpage

Considering the drawbacks of the template matching technique, we decided to use another technique for accurate eye center location estimation and tracking. This technique is based on image isophotes (intensity slices in the image). A full formulation of this algorithm is outside the scope of this manuscript, but the basic procedure is described next. For an in depth formulation refer to \cite{VALENTI&GEVERS}.

In Cartesian coordinates, the isophote curvature ($\kappa$) is:

\begin{equation}\label{eq:kappa}
\kappa = -\frac{L_{y}^{2}L_{xx} - 2L_{x}L_{xy}L_{y} + L_{x}^{2}L_{yy}}{(L_{x}^{2} + L_{y}^{2})^\frac{3}{2}}
\end{equation}
where $L_{x}$ and $L_{y}$ are the first-order derivatives of the luminance function $L(x,y)$ in the $x$ and $y$ dimension, respectively (\cite{DAM&ROMENY} and \cite{GINKEL&WEIJER&VLIET&VERBEEK}).

We can obtain the radius of the circle that generated the curvature of the isophote by reversing equation (\ref{eq:kappa}). The orientation of the radius can be estimated from the gradient but its direction will always point towards the highest change in the luminance. Since the sign of the isophote curvature depends on the intensity of the outer side of the curve (for a brighter outer side the sign is positive), by multiplying the gradient with the inverse of the isophote curvature, the duality of the isophote curvature helps in disambiguating the direction of the center. Then,

\begin{equation}
D(x,y) = -\frac{\{L_{x},L_{y}\}(L_{x}^{2} + L_{y}^{2})}{L_{y}^{2}L_{xx} - 2L_{x}LL_{xy}L_{y} + L_{x}^{2}L_{yy}}
\end{equation}
where $D(x,y)$ are the displacement vectors to the estimated position of the centers.

The curvedness, indicating how curved a shape is, was introduced as:

\begin{equation}
curvedness = \sqrt{L_{xx}^{2} + 2L_{xy}^{2} + L_{yy}^{2}}
\end{equation}

Since the isophote density is maximal around the edges of an object by selecting the parts of the isophotes where the curvedness is maximal, they will likely follow an object boundary and locally agree on the same center. Recalling that the sign of the isophote curvature depends on the intensity of the outer side of the curve, we observe that a negative sign indicates a change in the direction of the gradient (i.e. from brighter to darker areas). Regarding the specific task of cornea and iris location, it can be assumed that the sclera is brighter than the cornea and the iris, so we should ignore the votes in which the curvature is positive.

Then, a mean shift search window is initialized on the centermap, centered on the found center. The algorithm then iterates to converge to a region with maximal distribution. After some iteration, the center closest to the center of the search window is selected as the new eye center location.

\section{Tracking the Facial Features}\label{sec:features}

Since the second component of the visual gaze estimation pipeline uses both the information for the location of the facial features and the location of the eyes it is crucial that the features are tracked with the highest accuracy possible as well. In this early stage we use a simple approach for facial feature tracking based on the Lucas-Kanade method for optical flow estimation. The feature to track is selected in the first frame and it is tracked throughout the input sequence. Next the basic procedure is outlined, for a more in depth formulation of the algorithm refer to \cite{LUCAS&KANADE}.

\begin{figure}[htp!]
\begin{center}
\includegraphics[scale=1]{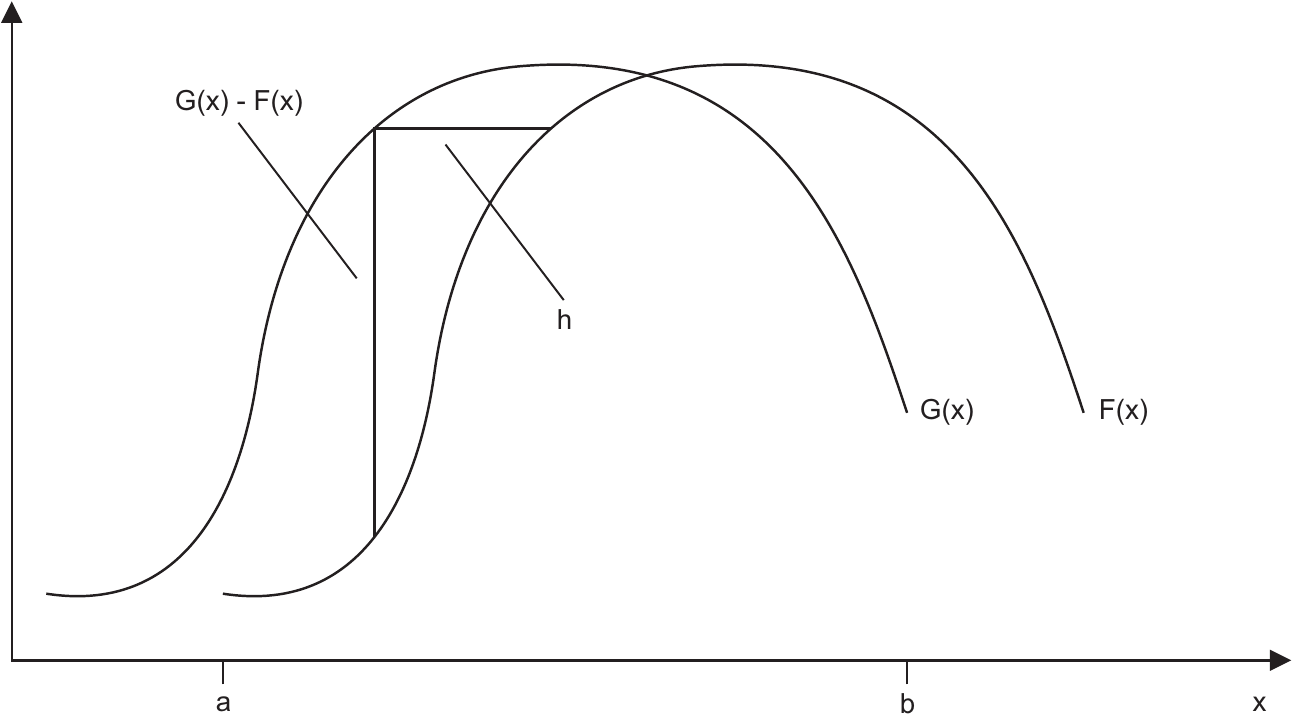}
\caption{One-dimensional image registration problem}
\label{fig:disparity}
\end{center}
\end{figure}

The translational image registration problem can be characterized as follows: we are given functions $F(x)$ and $G(x)$ which give the respective pixel values at each location $x$ in two images. We wish to find the disparity vector $h$ that minimizes some measure of difference between $F(x + h)$ and $G(x)$, for $x$ in some region of interest $R$. In the one-dimensional registration problem, we wish to find the horizontal disparity $h$ between two curves $F(x)$ and $G(x) = F(x + h)$ (see figure \ref{fig:disparity}).
\newpage

Like other optical flow estimation algorithms, the Lucas-Kanade method tries to determine the `movement' that occurred between two images. If we consider an one-dimensional image $F(x)$ that was translated by some $h$ to produce image $G(x)$, in order to find $h$, the algorithm first finds the derivative of $F(x)$ and then it assumes that the derivative can be approximated linearly for reasonably small $h$, so that,

\begin{equation}
G(x) = F(x + h) \approx F(x) + hF'(x)
\end{equation}

We can formulate the difference between $F(x + h)$ and $G(x)$ over the entire curve as an $\textbf{L}_{2}$ norm,

\begin{equation}
E = \displaystyle\sum_{x}[F(x + h) - G(x)]^{2}
\end{equation}

Then, to find the $h$ which minimizes this difference norm, we can set,

\begin{eqnarray}
0 & = & \frac{\partial E}{\partial h} \nonumber \\
& \approx & \frac{\partial}{\partial h}\displaystyle\sum_{x}[F(x) + hF'(x) - G(x)]^{2} \nonumber \\
& \approx & \displaystyle\sum_{x}2F'(x)[F(x) + hF'(x) - G(x)]
\end{eqnarray}

Using the above, we can deduce to,

\begin{equation}
h \approx \frac{\displaystyle\sum_{x}F'(x)[G(x) - F(x)]}{\displaystyle\sum_{x}F'(x)^{2}}
\end{equation}

This can be implemented in an iterative fashion,

\begin{equation}
\begin{array}{l}
h_{0} = 0, \\
h_{k + 1} = h_{k} + \frac{\displaystyle\sum_{x}F'(x + h_{k})[G(x) - F(x + h_{k})]}{\displaystyle\sum_{x}F'(x + h_{k})^{2}}
\end{array}
\end{equation}

Since at some points the assumption that $F(x)$ is linear holds better than at others, a weighting function is used to account for that fact. This function allows those points to play more significant role in determining $h$ and conversely, it gives less weight to points where the assumption of linearity (so $F(x)$ being close to 0) does not hold.

For the specific case of facial features tracking, we use the algorithm to obtain the coordinates of the facial features in some frame $b$ in the sequence, given the coordinates of the facial features in an earlier frame $a$, along with the image data of frames $a$ and $b$, allowing the tracking of the facial features.

So far we have described the algorithm used for detection and tracking of the location of the user's eyes and the algorithm used to define and track the location of the user's facial features in the sequence of images obtained through the web camera. The combination of these two algorithms is the body of the first component of the visual gaze estimation pipeline. In the second component of the pipeline this information is used to deduce the user's point of visual gaze on the computer screen.

\section{Calibration Methods}\label{sec:calibration}

\begin{figure}[htp!]
\begin{center}
\includegraphics[scale=1]{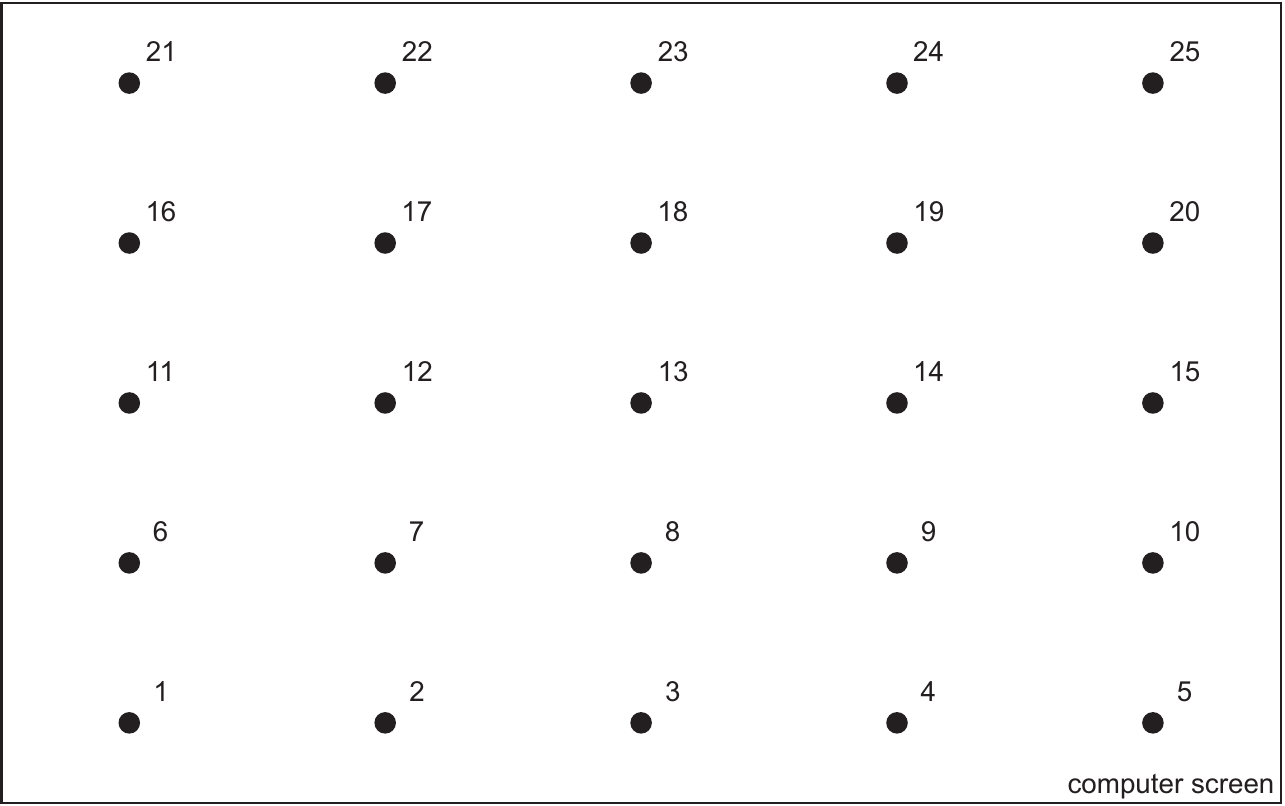}
\caption{Calibration template image}
\label{fig:calibrationTechniques}
\end{center}
\end{figure}

When a new image is captured through the web camera the first component of the visual gaze estimation pipeline constructs facial feature-pupil center vectors between the detected location of the user's facial features and the detected location of the user's eyes. Since the assumption in this scenario is that the user's head is static, the location of the facial features should not change too much in time, so they can play the role of anchor points. Then, the vectors collected during the calibration procedure will capture the offsets of the location of the user's eyes responsible for looking at different points on the computer screen; the precise mapping from facial feature-pupil center vectors to points on the computer screen is necessary for accurate visual gaze tracking system. The calibration procedure is done using a set of 25 points distributed evenly on the computer screen (see figure \ref{fig:calibrationTechniques}). The user is requested to draw points with the mouse in locations that are as near as possible to the locations in the template image.

It is inevitable to introduce errors in the system at this stage. Consider the following scenario: an user sitting in front of the computer screen and been requested to gaze at a certain point on it. We claim that even a human is not capable to say in pixel coordinates precision the coordinates of the point at which he/she is looking; a region around that point is the best we can get. A more elaborated argument on this observation is offered later in the manuscript (see section \ref{sec:experimentsConclusions}) by close examination of the physiological limitations imposed by the structure of the human eye and the process of visual perception in the human beings. We choose this type of calibration procedure because when the user is tracking the mouse cursor movement with his/her eyes and is clicking at the desired position, we ensure to record as close as possible, the user's eye saccade that is paired with the point of interest on the computer screen. Finally, research done on the choice of the cursor shape for calibration purposes shows that the cursor's shape has an impact on the precision with which a human looks at it. Since our goal is to build a calibration free system where these observations and choices will not play any role we will not elaborate further. Once the user clicks on the computer screen, the facial feature-pupil center vector is recorded together with the location of the point on the computer screen.

After the calibration procedure is done, several models for mapping are used: 2D mapping with interpolation model that uses 2 points (figure \ref{fig:calibrationTechniques}, points: 21 and 5 or 1 and 25), linear model that uses 5 points (figure \ref{fig:calibrationTechniques}, points: 21, 25, 13, 1 and 5), and second order model that uses 9 or 25 points (figure \ref{fig:calibrationTechniques}, points: 21, 23, 25, 11, 13, 15, 1, 3 and 5 or all). The details on the model selection and the mathematics behind it are given next.

\section{Estimating the Point of Visual Gaze}\label{sec:solution}

The problem we are facing is to compute the point of visual gaze in screen coordinates given the current facial feature-pupil center vector in camera coordinates. The facial feature-pupil center vector is constructed upon the current location of the user's eye and the location of the user's facial feature. We can see this problem, as like there is an unknown function that maps our input (facial feature-pupil center vector) to specific output (point on the computer screen). Furthermore, this problem can be seen as a simple regression problem, namely, suppose we observe a real-valued input variable $x$ (facial feature-pupil center vector) and we wish to use this observation to predict the value of a real-valued target variable $t$ (point on the computer screen). We would like to learn the model (function) that is able to generate the desired output given the input.

Suppose we are given a training set (constructed during the calibration procedure) comprising $N$ observations of $x$, together with the corresponding observations of the values of $t$. Our goal is to exploit the training set in order to make predictions of the value $\Hat{t}$ of the target variable for some new value $\Hat{x}$ of the input variable. If we consider a simple approach based on curve fitting then we want to fit the data using a polynomial function of the form:

\begin{equation}
y(x,\textbf{w}) = w_{0} + w_{1}x + w_{2}x^{2} + ... + w_{M}x^{M} = \displaystyle\sum_{j=0}^{M}w_{j}x^{j}
\end{equation}
where $M$ is the order of the polynomial.

The values of the coefficients will be determined by fitting the polynomial to the training data. This can be done by minimizing an error function that measures the misfit between the function $y(x,\textbf{w})$, for any given value of $\textbf{w}$, and the training set data points. One simple choice of error function, which is widely used, is given by the sum of the squares of the errors between the predictions $y(x_{n},\textbf{w})$ for each data point $x_{n}$ and the corresponding target values $t_{n}$, so that we minimize

\begin{equation}
E(\textbf{w}) = \frac{1}{2}\displaystyle\sum_{n=1}^{N}\{y(x_{n},\textbf{w}) - t_{n}\}^{2}
\end{equation}

The error function is a nonnegative quantity that would be zero if, and only if, the function $y(x,\textbf{w})$ were to pass exactly through each training data point. We can solve the curve fitting problem by choosing the value of $\textbf{w}$ for which $E(\textbf{w})$ is as small as possible.

There remains the problem of choosing the order $M$ of the polynomial or so-called model selection. The models chosen for fitting our data are discussed next,

\begin{itemize}\label{list:models}
\item{\textbf{2D mapping with interpolation model} takes into account 2 calibration points (figure \ref{fig:calibrationTechniques}, points: 21 and 5 or 1 and 25). The mapping is computed using linear interpolation as follows:
\begin{eqnarray}
s_{x} = s_{x1} + \frac{x - x_{1}}{x_{2} - x_{1}}(s_{x2} - s_{x1}) \nonumber \\
s_{y} = s_{y1} + \frac{y - y_{1}}{y_{2} - y_{1}}(s_{y2} - s_{y1})
\end{eqnarray}

For example, suppose the screen coordinates and the facial feature-pupil center vectors used for calibration in points P1 and P2 are respectively $(s_{x1},s_{y1}$), $(x_{1},y_{1})$ and $(s_{x2},s_{y2})$, $(x_{2},y_{2})$. Then after the measurement of the current facial feature-pupil center vector $(x,y)$ is taken, the screen coordinates is computed using the above equations.}
\item{\textbf{linear model} takes into account 5 calibration points (figure \ref{fig:calibrationTechniques}, points: 1, 5, 13, 21, 25). The mapping between facial feature-pupil center vectors and screen coordinates is done using the following equations:
\begin{eqnarray}
s_{x} = a_{0} + a_{1}x \nonumber \\
s_{y} = b_{0} + b_{1}y
\end{eqnarray}
where $(s_{x},s_{y})$ are screen coordinates and $(x,y)$ is the facial feature-pupil center vector. The coefficients $a_{0},a_{1}$ and $b_{0},b_{1}$ are the unknowns and can be found using least squares approximation (appendix \ref{chapter:AppendixB}).}
\item{\textbf{second order model} takes into account a set of 9 (figure \ref{fig:calibrationTechniques}, points: 1, 3, 5, 11, 13, 15, 21, 25, 23) or a set of 25 (figure \ref{fig:calibrationTechniques}, all points) calibration points. The mapping between facial feature-pupil center vectors and screen coordinates is done using the following equations:
\begin{eqnarray}
s_{x} = a_{0} + a_{1}x + a_{2}y + a_{3}xy + a_{4}x^{2} + a_{5}y^{2} \nonumber \\
s_{y} = b_{0} + b_{1}x + b_{2}y + b_{3}xy + b_{4}x^{2} + b_{5}y^{2}
\end{eqnarray}
where $(s_{x},s_{y})$ are screen coordinates and $(x,y)$ is the facial feature-pupil center vector. The coefficients $a_{0},a_{1},a_{2},a_{3},a_{4},a_{5}$ and $b_{0},b_{1},b_{2},b_{3},b_{4},b_{5}$ are the unknowns and can be found using least squares approximation (appendix \ref{chapter:AppendixB}).}
\end{itemize}
\cleardoublepage

\chapter{Natural Head Movement}\label{chapter:NHM}

\dropping{2}{I}\newline n this chapter we propose a visual eye-gaze tracker that estimates and tracks the user's point of visual gaze on the computer screen under natural head movement, with the usage of a single low resolution web camera. The tracker implements two procedures that are suitable for different situations - in the rest of the manuscript the first one is regarded as \textit{2.5D} and it is suitable for situations when the user's head does not move too much, while the second one, regarded as \textit{3D} in the rest of the manuscript, is suitable when the user's head performs large movements.

This tracker can be classified as non-intrusive, 3D, and appearance based. As the tracker described in chapter \ref{chapter:SH}, it is not based on PCCR (Pupil Center-Corneal Reflection) technique, but solely on image analysis. The tracker is again composed of four steps (two steps for each component of the visual gaze estimation pipeline; see figure \ref{fig:gazePipeline}). In the first component of the pipeline we combine an algorithm for detection and tracking of the location of the user's eyes with an algorithm for detection and tracking of the position of the user's head; the origin of the user's head plays the role of the anchor points (facial features in the 2D tracker described in chapter \ref{chapter:SH}) so the eyes displacement vectors that are used in the second component of the pipeline are constructed upon the location of the origin of the head and the location of the eyes. The second component of the pipeline combines a calibration procedure that gives correspondences between head origin-pupil center vectors and points on the computer screen (train data), and a method (see appendix \ref{chapter:AppendixB}) to approximate a solution of overdetermined system of equations for future visual gaze estimation.

Huge amount of research has been done on possible solutions of the problem of visual gaze tracking under natural head movement in this settings, and to the best of our knowledge there is no tracker in the literature similar to the proposed one. The main idea behind the proposed tracker is the observation that the scenario in which the user is restricted in head movement is a special case of the scenario in which the user is granted with the freedom of natural head movement. The fact that we are using a single camera bounds us in two dimensions (no depth can be recovered). Furthermore, the definition of the point of visual gaze (see figure \ref{fig:gazeDefinitionIntro}) infers 3D modeling of the eye and the scene set-up in order to compute the intersection of the visual axis of the eye with the computer screen. In chapter \ref{chapter:SH} we described a procedure that omits this 3D modeling by introducing a calibration step where correspondences between facial feature-pupil center vectors and points on the computer screen are constructed. The drawback of the calibration step is the fact that these correspondences refer to specific position of the head in 3D space (the position at which that calibration is done). In case the position of the head differs too much from the original position, the mapping obtained in the calibration step fails resulting in erroneous estimate for the point of visual gaze.

In essence the calibration procedure is not affected by the position of the head; for example, if we perform calibration as described in section \ref{sec:calibration} when the head is randomly rotated and/or translated in space, the 2D tracker will perform with the same accuracy as when calibrated with frontal head (assuming that in both cases the position of the head stays as close as possible to the position at which the calibration is done). Then, if we could re-calibrate the tracker each time the head is rotated and/or translated, we could exploit the simplicity of the solution in two dimensions (omitting again the need of 3D modeling) while removing the restriction on the user to keep his/her head as static as possible. In the rest of the chapter we describe how to transform the 3D problem into many 2D special cases and then to estimate the point of visual gaze in two dimensions overcoming the restriction on the tracker to use a single camera.

\section{Proposed Method}

\begin{figure}[htp!]
\begin{center}
\includegraphics[scale=1]{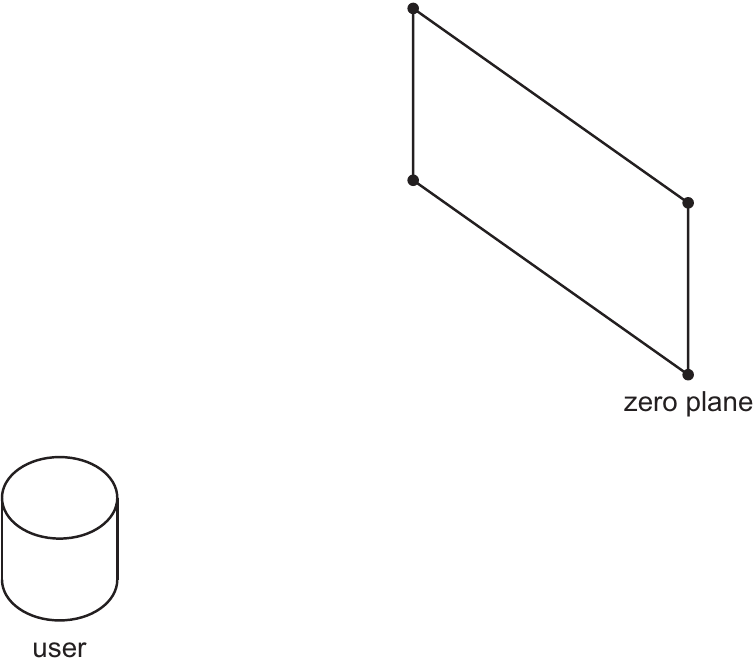}
\caption{The scene set-up in 3D}
\label{fig:scenario3D1}
\end{center}
\end{figure}

The 2D tracker estimates the point of visual gaze according to the user's facial feature-pupil center vector in the current frame. The tracker feeds the learned model (function) during the calibration step with the current facial feature-pupil center vector and then the learned model calculates the output - the user's point of visual gaze in the current frame. Then, as mentioned previously, the accuracy is related to the position of the head in the current frame; the more the head position differs from the original position (at which the calibration is done) the poorer the accuracy of the estimated point of visual gaze (because the mapping becomes inconsistent). Then, in figure \ref{fig:scenario3D1} we start examining the problem in three dimensions; the figure illustrates the same scene set-up (as in chapter \ref{chapter:SH}) in 3D space where the cylinder represents the head position and the computer screen is regarded as the \textit{zero plane}.
\newpage

Our idea is to re-calibrate each time the user moves his/her head. Recalling that the calibration step takes into account the head origin-pupil center vectors and the corresponding points on the zero plane, we have two options for re-calibration. We can keep the location of the calibration points on the zero plane static and correct the head origin-pupil center vectors that correspond to these points once the head is moved (a proposition in this direction is given in section \ref{sec:improvements}). This means that each time the head moves we need to request the user to perform new calibration. Although, this is the correct choice and it ensures an accurate estimation of the point of visual gaze, it is extremely intrusive. 

\begin{figure}[htp!]
\begin{center}
\includegraphics[scale=1]{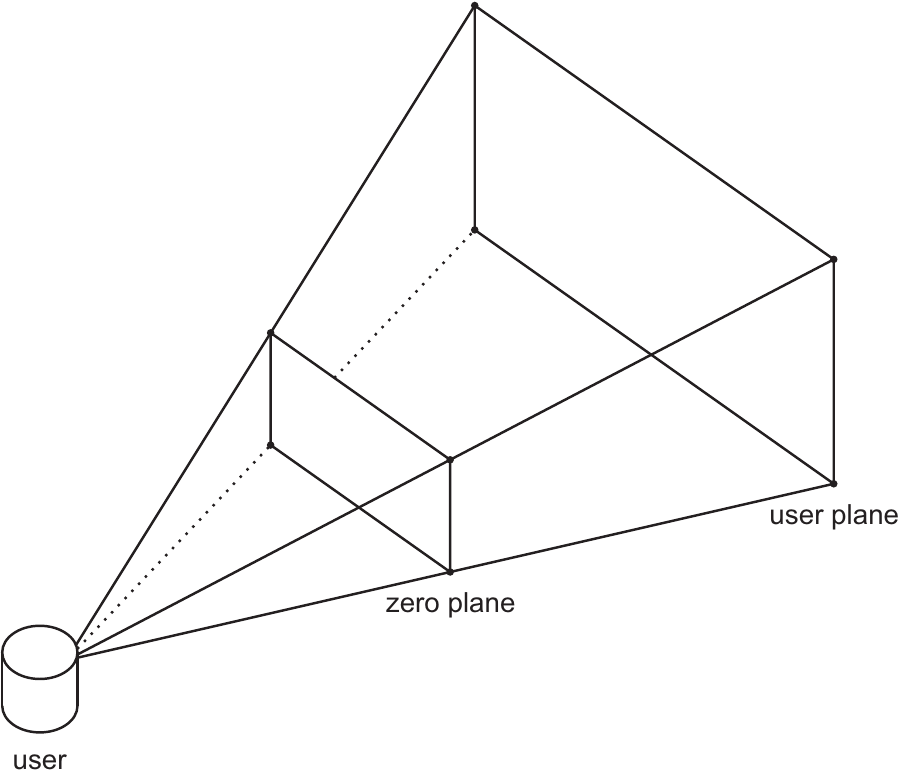}
\caption{User plane attached to the user's head (construction)}
\label{fig:scenario3D4}
\end{center}
\end{figure}

\begin{figure}[htp!]
\begin{center}
\includegraphics[scale=1]{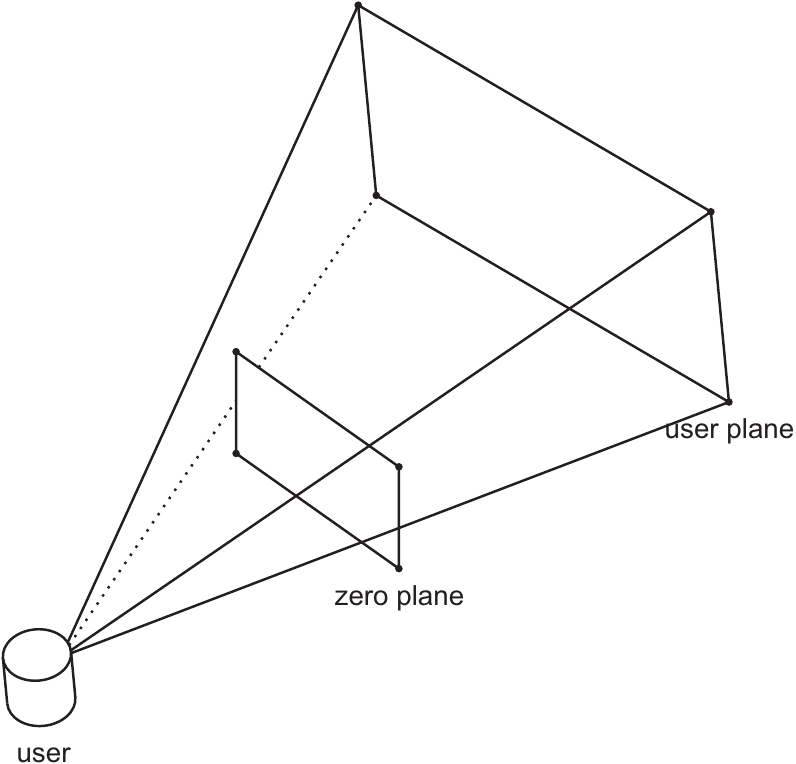}
\caption{User plane attached to the user's head (movement)}
\label{fig:scenario3D5}
\end{center}
\end{figure}

Realizing that this approach is not acceptable we are left with one choice - we can keep the set of head origin-pupil center vectors obtained during the calibration step static and correct the location of the calibration points once the head is moved. Evidently, this is a compromise from which the accuracy suffers, but we believe that it is a step in the right direction. Consider the following scenario - the user sits in front of the computer screen and rotates his/her head so its origin is projected to the vertical center of the rightmost part of the screen, meaning that the user is looking somewhere in this part of the screen. If we translate all calibration points to this part of the screen (actually, half of the points will be in the right outer part of the screen) and use the previously obtained head origin-pupil center vectors we can re-calibrate and achieve reasonable accuracy for the estimate of the point of visual gaze (assuming that the user is not behaving abnormal, say, rotates his/her head as described and looks in the leftmost part of the computer screen). We start this procedure by defining and constructing an \textit{user plane} that is attached to the head. Constructing the user plane in this way ensures that when the user moves his/her head, the user plane will move according to the head movement (see figure \ref{fig:scenario3D4} and figure \ref{fig:scenario3D5}).
\newpage

\vskip 2 truecm

During the calibration procedure the user is requested to draw the calibration points (see the calibration template image; figure \ref{fig:calibrationTechniques}) on the computer screen. In the 3D model of the scene set-up the computer screen is regarded as the zero plane, so during the calibration we populate the zero plane with the user defined calibration points. Since the zero plane is static in the 3D model of the scene but we want to capture the head movement we need to translate the calibration points from the zero plane to the user plane. We do this by tracing a ray starting at the head origin (cylinder's origin) to a calibration point on the zero plane. Then, we calculate the intersection point of that ray with the user plane. This procedure is repeated for each calibration point, so a set of \textit{user points} is defined during the calibration procedure and that set of user points is used to correct the calibration when the user moves his/her head. Refer to appendix \ref{chapter:appendixC} for detailed description on the mathematics behind line-plane intersection in 3D space.

\begin{figure}[htp!]
\begin{center}
\includegraphics[scale=1]{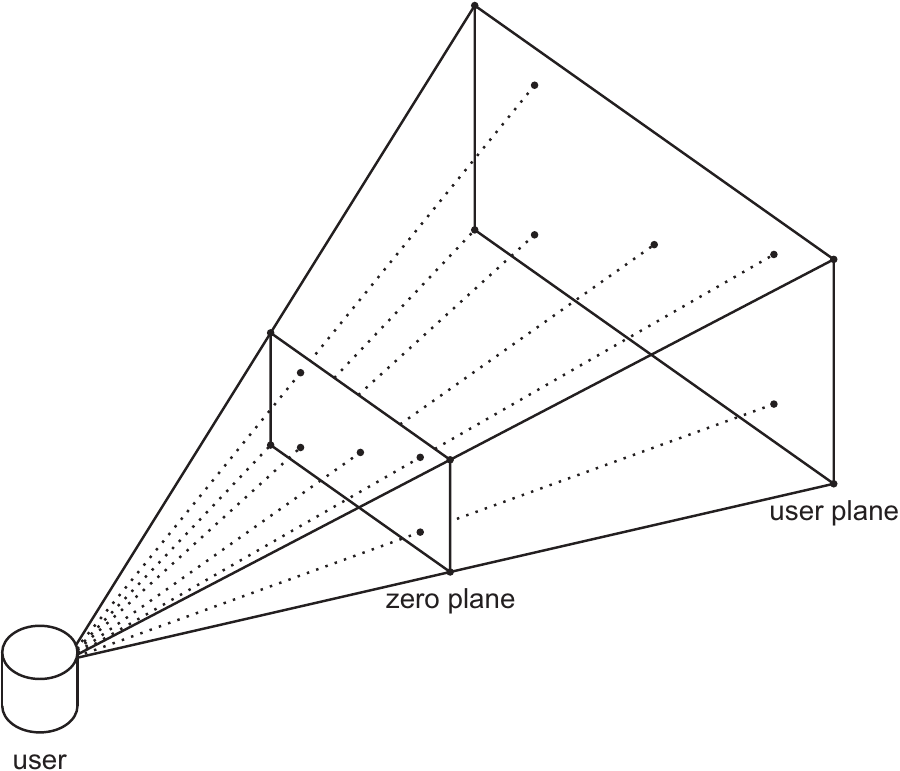}
\caption{User points and their intersections with the zero plane (construction)}
\label{fig:scenario3D2}
\end{center}
\end{figure}

\begin{figure}[htp!]
\begin{center}
\includegraphics[scale=1]{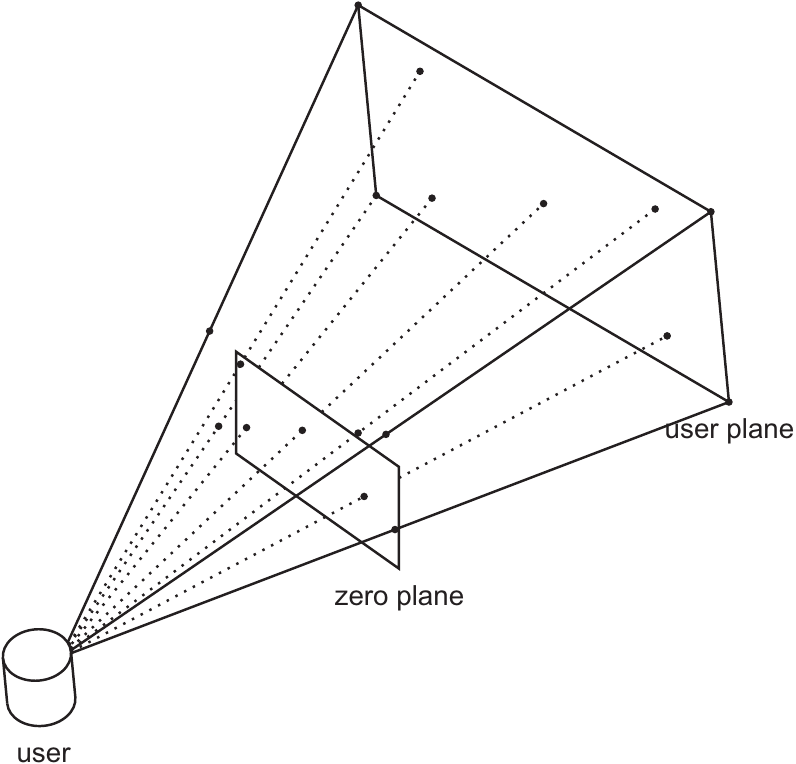}
\caption{User points and their intersections with the zero plane (movement)}
\label{fig:scenario3D3}
\end{center}
\end{figure}

Due to the properties of the user plane - it is attached to the head, when the user moves his/her head the obtained user points will change their location based on the head movement. At this point, we can trace back a ray starting at an user point to the head origin (cylinder's origin). Then, we calculate the intersection point of that ray with the zero plane. If we repeat this procedure for each user point, a set of new calibration points on the zero plane can be constructed each time the user moves his/her head. Having the new set of calibration points and the static head origin-pupil center vectors, we can learn a new model (function) based on the new the training set of data points, and feed that model with the current head origin-pupil center vector (solve for the point of visual gaze in the current frame). Figure \ref{fig:scenario3D2} illustrates the idea of constructing the set of user points during the calibration procedure and figure \ref{fig:scenario3D3} illustrates a movement of the head and the new set of intersection points (calibration points) used for re-calibration.
\newpage

\vskip 2 truecm

One may wonder why we are not using directly the zero plane but instead we are constructing an user plane and calculating intersection points with the zero plane. Close observation on the path that the user plane follows when the user moves his/her head shows that it is an arc-like path. This means that if we translate directly the calibration points (on the zero plane) their possible locations will be distributed on the surface of the inner part of a hemisphere (which origin is the origin of the cylinder and radius the distance from the cylinder to the zero plane). Assuming that the computer screen is planar, if we choose that option we will introduce errors in the location of the set of calibration points. This is the reason to construct an user plane and calculate the rays intersections with the zero plane, effectively keeping the planarity of the set of calibration points.

Finally, it is easy to see that the static head scenario described in chapter \ref{chapter:SH} is a special case of the scenario when the head can perform movements. Treating every movement of the head as a special case and redefining the set of calibration points based on that movement gives us the potential to exploit the simplicity of the solution for the point of visual gaze in two dimensions. Furthermore, this approach helps us to overcome the restriction of a single camera usage and the in-feasibility of 3D modeling of the eye and the scene. Unfortunately, there is a drawback associated with this approach, as reasoned before, new errors (calibration wise) are introduced in the system at this step (because we keep the head origin-pupil center vectors static).

\section{Implementation}\label{sec:implementation}

For this tracker we use a cylindrical head pose tracker algorithm based on the Lukas-Kanade optical flow method \cite{XIAO&KANADE&COHN}. The integration of the cylindrical head pose tracker and the isophote based eye tracker (described in section \ref{sec:eyes}) is outside the scope of this manuscript, refer to \cite{VALENTI&YUCEL&GEVERS} for detailed description. Next we give a short description of the initialization step of the cylindrical head pose tracker.

The first time a frontal face is detected in the image sequence, the initial parameters of the cylinder and its initial transformation matrix are computed: the size of the face is used to initialize the cylinder parameters and the pose $p = [w_{x}, w_{y}, w_{z}, t_{x}, t_{y}, t_{z}]$, based on anthropometric values \cite{DODGSON}, \cite{GORDON&BRADTMILLER&CHURCHILL&CLAUSER&MCCONVILLE&TEBBETTS&WALKER}, where $w_{x}$, $w_{y}$, and $w_{z}$ are the rotation parameters and $t_{x}$, $t_{y}$, $t_{z}$ are the translation parameters. The location of the eyes is detected in the face region and it is used to give a better estimate of the $t_{x}$ and $t_{y}$. The depth, $t_{z}$, is adjusted by using the distance between the detected eyes, $d$. The detected face is assumed to be frontal, so the initial pitch ($w_{x}$) and yaw ($w_{y}$) angles are assumed to be null, and the roll angle ($w_{z}$) is initialized by the relative position of the eyes.

For the facial feature location we use the location of the origin of the user's head (the origin of the cylinder), so we do not need to define it manually anymore (as in the tracker described in chapter \ref{chapter:SH}). The head origin-pupil center vectors are obtained in pose normalized coordinates. We can exploit this fact by constructing a set of pose normalized head origin-pupil center vectors which can be used for future calibrations. This can be achieved by asking the user to start the tracker once his/her head is at certain position relative to the computer screen. This will ensure that we can use a set of predefined head origin-pupil center vectors and a set of predefined calibration points on the screen; this in turn leads to calibration free system (further elaborated in section \ref{sec:improvements}).

Since we reduced the problem back to two dimensions by treating the head movement in space as a special case in two dimensions we can use the same calibration method as described in section \ref{sec:calibration} and the same approach for estimating the point of visual gaze as described in section \ref{sec:solution}.
\cleardoublepage

\chapter{Experiments}\label{chapter:experiments}

\begin{figure}[htp!]
\begin{center}
\subfigure[center]{
\includegraphics[scale=1]{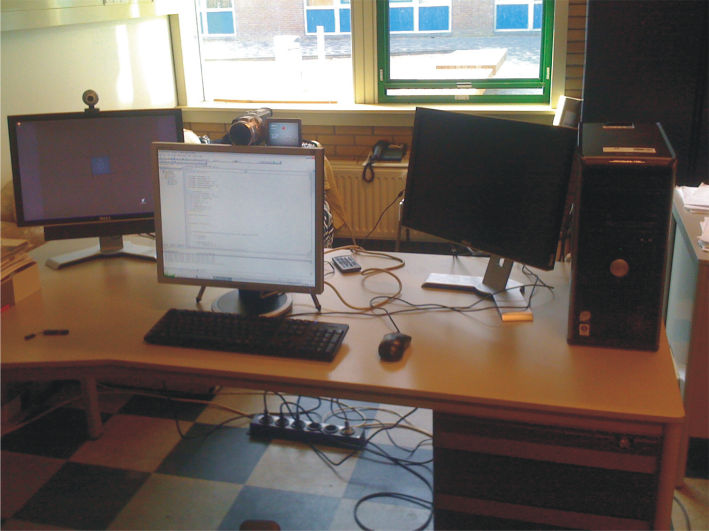}}
\subfigure[right]{
\includegraphics[scale=1]{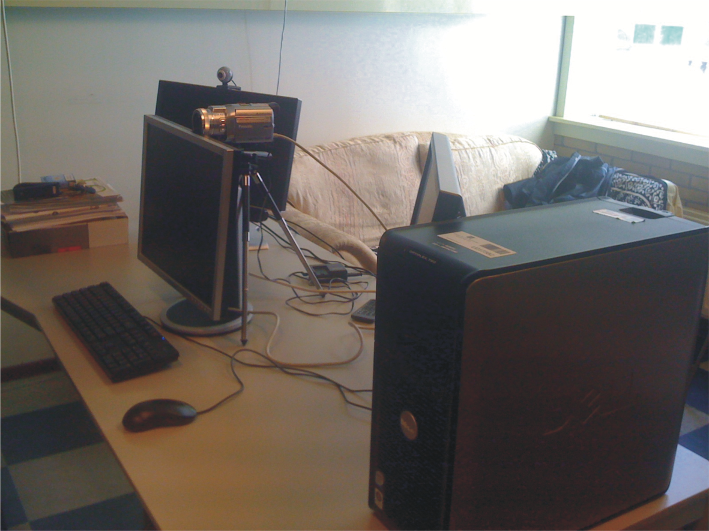}}
\caption{The set-up for the experiments}
\label{fig:experimentSetUp}
\end{center}
\end{figure}

\dropping{2}{T}\newline his chapter describes the experiments performed to evaluate the proposed visual eye-gaze trackers. For evaluation purpose, we designed and conducted two different experiments. The goal of the first experiment (called \textit{static}) is to test the accuracy of the proposed systems when the test subject's head is static (see section \ref{sec:static}). The goal of the second experiment (called \textit{dynamic}) is to test the accuracy of the proposed trackers in dynamic settings, i.e., the subject is granted with the freedom of natural head movement (see section \ref{sec:dynamic}). The chapter discusses the choices made during the experiments' design and insights regarding possible origins of errors. Then, we describe a third experiment conducted based on these observations (see section \ref{sec:experimentNHM}). Figure \ref{fig:experimentSetUp} illustrates the set-up for the experiments.

\begin{figure}[htp!]
\begin{center}
\includegraphics[scale=1]{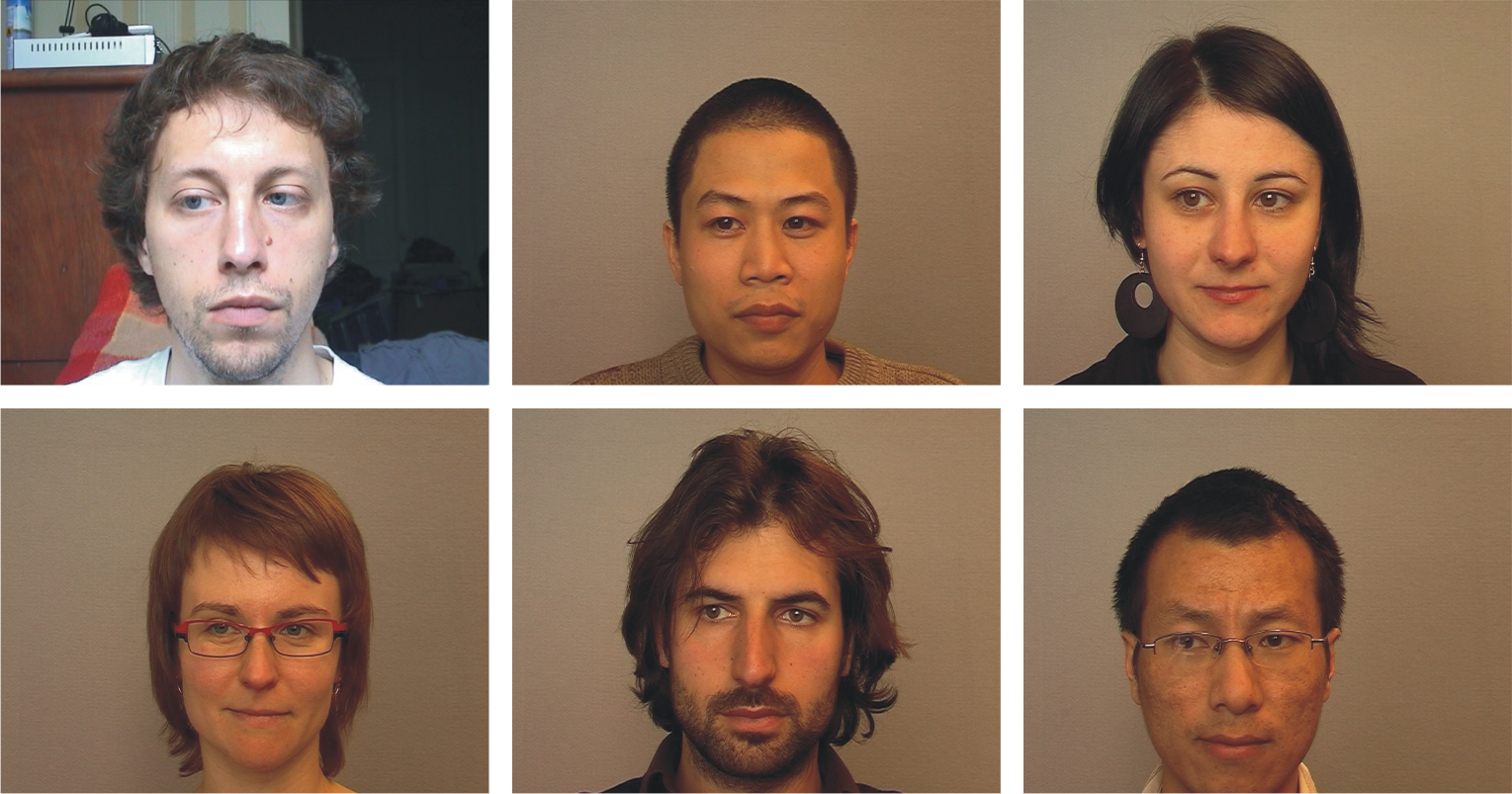}
\caption{Test subjects}
\label{fig:testSubjects}
\end{center}
\end{figure}

During the experiments, a dataset is collected that is highly heterogeneous; it includes both male and female participants, different ethnicity, with and without glasses; furthermore, it includes different illumination conditions. In order to conform with the usability requirements outlined in section \ref{sec:problem} the experiments are performed without the use of chin-rests, so even the experiment that we call static includes subtle movements of the subjects' head. Indeed, close observation of the video footages collected during the static experiment, shows that the subjects tend to rotate their heads towards the point of interest on the computer screen. Although, this rotation is not extreme, it is a perfect example of a natural head movement in front of the screen. Figure \ref{fig:testSubjects} shows some examples of the subjects in the dataset.

Some of the settings of the experiments are that the subject's head is at distance of roughly $750mm$ from the computer screen. The camera coordinate system is defined in the middle of the streamed pictures/video and the origin of the subject's head is as close as possible to the origin of the camera coordinate system. The resolution of the captured images is 720 $\times$ 576 pixels and the resolution of the computer screen is 1280 $\times$ 1024 pixels. The point displayed on the screen is a circle with radius of 25 pixels - resulting in a blob on the screen that has dimensions of approximately 50 $\times$ 50 pixels.

\section{No Head Movement}\label{sec:static}

\begin{figure}[htp!]
\begin{center}
\includegraphics[scale=1]{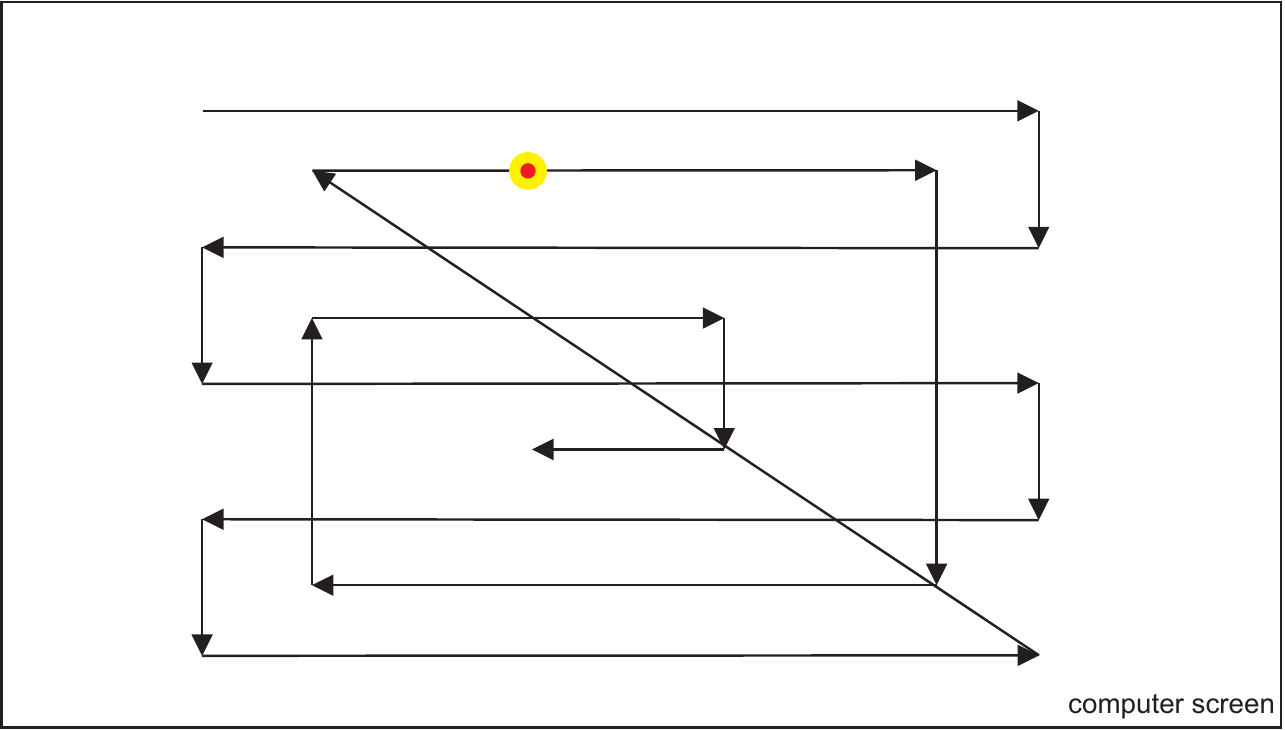}
\caption{Static experiment screenshot}
\label{fig:experimentStatic}
\end{center}
\end{figure}

We call this experiment static because the test subject is asked to keep his/her head as close as possible to the starting position (no head movement). The duration of the experiment is about 2.5 minutes, so it is important that the experiment starts only when the subject feels comfortable enough with his/her current pose, ensuring collection of meaningful data. Once the subject feels comfortable enough, the experiment starts. The goal of the subject is to follow only with his/her eyes a moving point that follows a predefined path on the computer screen (see figure \ref{fig:experimentStatic}). The point's animation starts in the middle of the screen and moves up and left to a location close to the top left corner. In this initial frames no data is collected - the purpose of the initial frames of the animation is to give time to the subject to get used with the point's movement and to have time to start following the animation as accurate as possible. Data collection begins once the point reaches a location close to the top left corner of the screen (the beginning of the path in figure \ref{fig:experimentStatic}). The arrows in figure \ref{fig:experimentStatic} show the step by step movement of the point. The restrictions in the experiment are that the subject must keep his/her head as static as possible and must avoid blinking.

At each time step (25 frames per second) an image of the test subject is recorded while he/she is gazing at the point. The dataset for each subject is composed of a set of pictures of him/her gazing at different locations on the computer screen and the coordinates of the points corresponding to these locations.

\section{Extreme Head Movement}\label{sec:dynamic}

\begin{figure}[htp!]
\begin{center}
\includegraphics[scale=1]{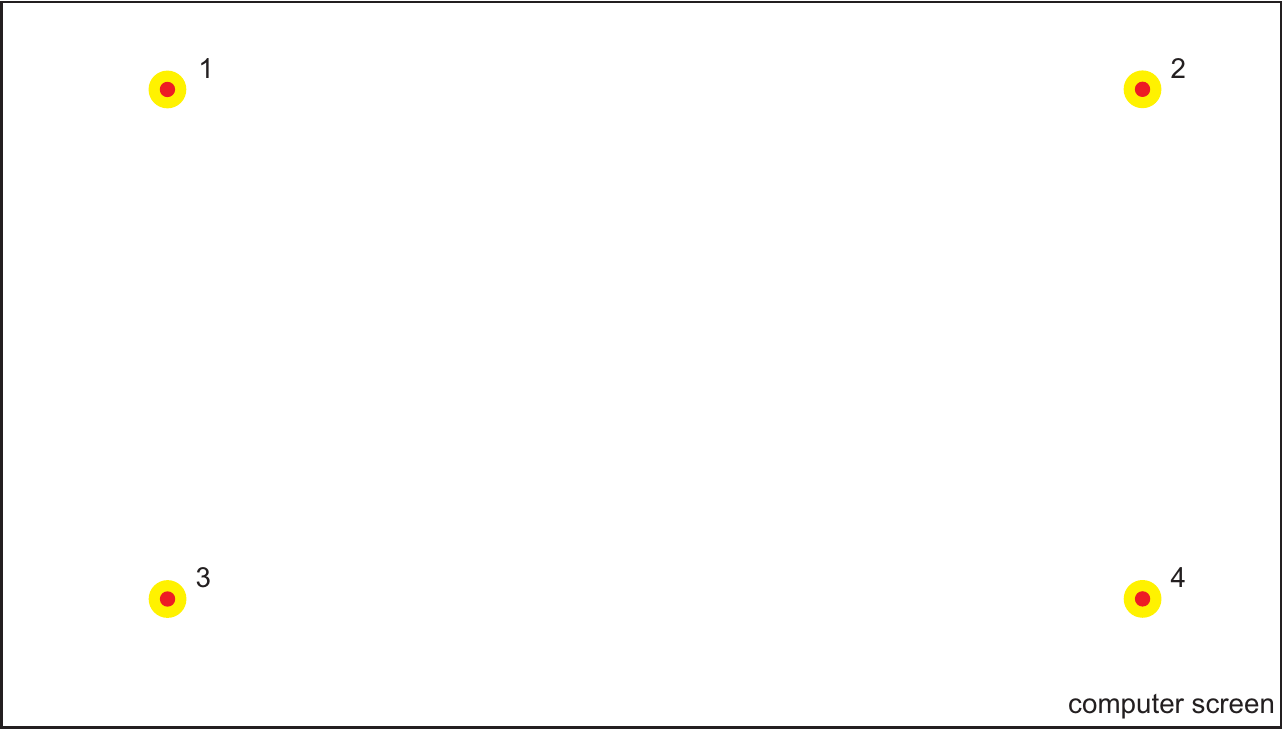}
\caption{Dynamic experiment screenshot}
\label{fig:experimentDynamic}
\end{center}
\end{figure}

The second experiment is called dynamic because the test subject is asked to move his/her head in a specific way. The duration of the experiment is about 15 seconds. The subject is requested to gaze with his/her eyes at a static point on the computer screen (see figure \ref{fig:experimentDynamic}) and move his/her head around while still looking at the point. The point is displayed at certain locations on the screen for about 4 seconds each time. When the point is displayed on the screen the subject is asked to look at it and then to rotate his/her head towards the point's location. When the desired head position is reached the subject is asked to say `yes' and `no' with his/her head while still gazing at the point. The data collection begins in the moment the first point is displayed. The order in which the point is displayed is visually described by the numbers in figure \ref{fig:experimentDynamic}. The restriction in the experiment is that the subject must avoid blinking.

At each time step (25 frames per second) an image of the test subject is recorded while he/she is gazing at the static point and moves his/her head. The data for each subject is a set of pictures of him/her gazing at the four different locations on the computer screen while moving his/her head, and the coordinates of these locations.

\section{Observations}\label{sec:experimentsConclusions}

There are several important observations drawn from the conducted experiments,

\begin{itemize}
\item{the camera resolution in both experiments is not much higher than the resolution of an ordinary web camera. Indeed, it is possible to buy a web camera with higher resolution than the used one and it will be on acceptable price;}
\item{we observed that the test subjects are more concentrated at the beginning of the experiments than at the end, resulting in more errors introduced with the time elapsed;}
\item{in order to simulate a real life situation, no chin-rests are used - this introduces errors in trackers and decreases the accuracy of the estimate for the point of visual gaze. Although, this is undesirable, this type of test gives a glance on the accuracy of the systems when used in a normal, uncontrolled environment. Besides the screen resolution, the camera resolution, and the point's dimensions, there are no variables in the experiments that are closely controlled and can be taken as constants. Although, this will introduce even more errors in the visual gaze estimation, there is an upside of conducting the experiments in such way - now it is easy to argue that there is no special hardware and settings involved in the final result. This makes the system re-usable in any condition and by anyone and ensures that the results are as close as possible to the results that any person can achieve in any settings;}
\item{during the test phase it became clear that it is important to define what a natural head movement is, when the user is situated at approximately $750mm$ from the computer screen. There are important insights regarding the design of the dynamic experiment that can follow by close investigation of this definition.}
\end{itemize}

The later observation and the fact that so far we had been mainly concerned with the \textit{computer vision} part of the problem and we had never put under consideration its physiological and psychological aspects inspired the following reasoning. We want to define what a natural head movement is in front of the computer screen in this settings and what/where is the maximal accuracy of the visual gaze.

\begin{figure}[htp!]
\begin{center}
\subfigure[horizontal]{
\includegraphics[scale=1]{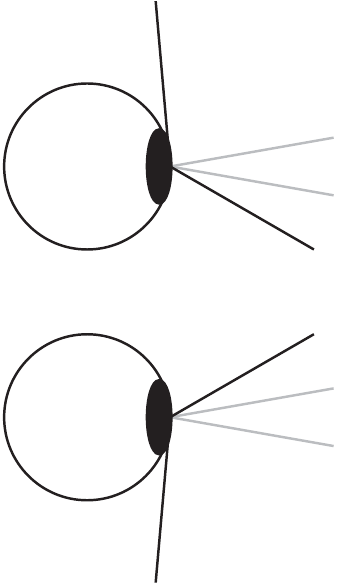}}
\subfigure[vertical]{
\includegraphics[scale=1]{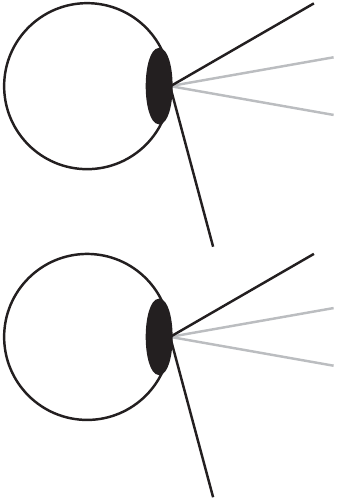}}
\caption{Human visual field}
\label{fig:humanFOV}
\end{center}
\end{figure}

It seems that there is not much research in the literature regarding what we could call a natural head movement in front of the computer screen. We want to define it so we start our investigation from the visual field of the human. The approximate field of view of the human eye is $95^{\circ}$ out, $75^{\circ}$ down, $60^{\circ}$ in, $60^{\circ}$ up. Figure \ref{fig:humanFOV} illustrates the visual field of the human for both eyes simultaneously. The black lines represent the whole field of view while the gray lines represent the $10^{\circ}$ around the optical axis of the eye. About $12^{\circ}$ - $15^{\circ}$ temporal and $1.5^{\circ}$ below the horizontal is the optic nerve or the blind spot which is roughly $7.5^{\circ}$ in height and $5.5^{\circ}$ in width \cite{MIL-STD-1472F}. The blind spot is the place in the visual field that corresponds to the lack of light-detecting photoreceptor cells on the optic disc of the retina where the optic nerve passes through it. Since there are no cells to detect light on the optic disc, a part of the field of vision is not perceived. The brain fills in the spot with surrounding details and with information from the other eye, so the blind spot is not normally perceived.
\newpage

\begin{figure}[htp!]
\begin{center}
\includegraphics[scale=1]{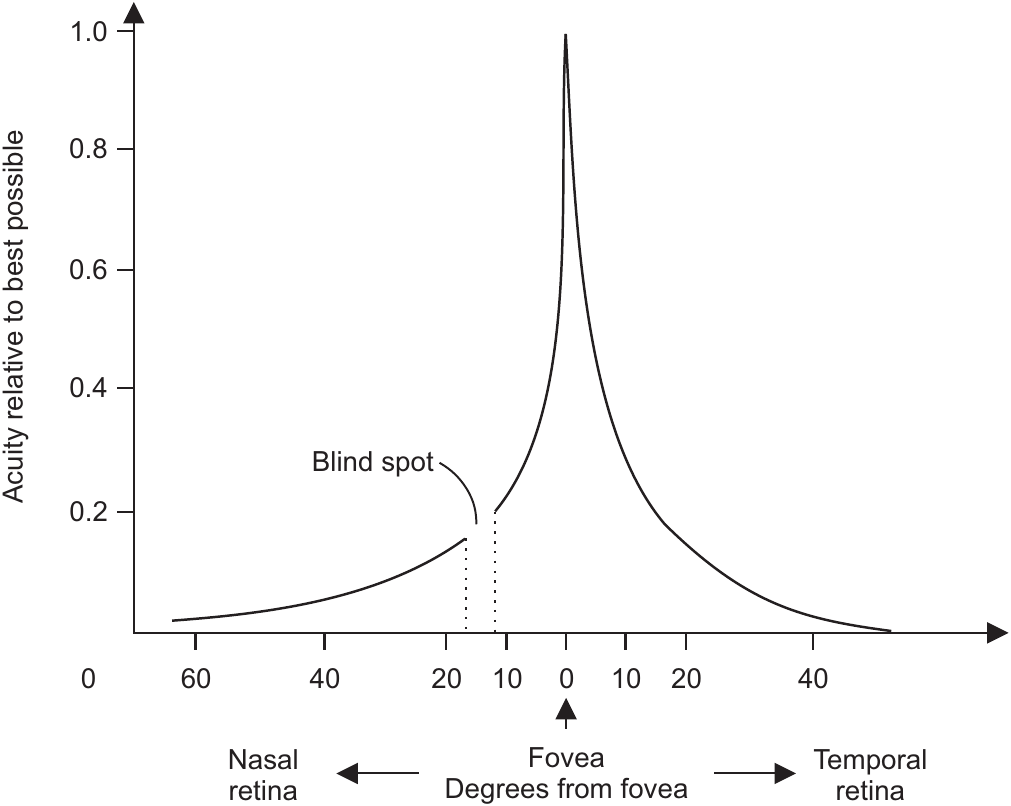}
\caption{Relative acuity of the left human eye (horizontal section) in degrees from the fovea}
\label{fig:acuityHumanEye}
\end{center}
\end{figure}

The fovea is responsible for the sharp central vision (also called foveal vision), which is necessary in humans for reading, watching television or movies, driving, and any activity where visual detail is of primary importance. According to \cite{HUNZIKER} acuity of the human eye drops drastically at the $10^{\circ}$ from the fovea. Figure \ref{fig:acuityHumanEye} shows the relative acuity of the left human eye (horizontal section).

There is even further complication when thinking of what the visual perception is. The capability of the brain to fill places in the visual field where there is low or no information comes in support of the argument that even a human is not able to pinpoint in pixel accuracy the point of interest on the computer screen. Further, the definition of the visual acuity itself takes into consideration the brain interpretation ability: ``visual acuity is acuteness or clearness of vision, especially for vision, which is dependent on the sharpness of the retinal focus within the eye and the sensitivity of the interpretative faculty of the brain''. From psychological point of view, the major problem in visual perception is that what people see is not simply a translation of retinal stimuli (i.e., the image on the retina). Thus, people interested in perception have long struggled to explain what visual processing does to create what we actually see. It turns out that we might \textit{believe} that we look at certain point when asked while the perception is constructed upon interplays between past experiences, including one's culture, and the interpretation of the perceived.

\begin{figure}[htp!]
\begin{center}
\includegraphics[scale=1]{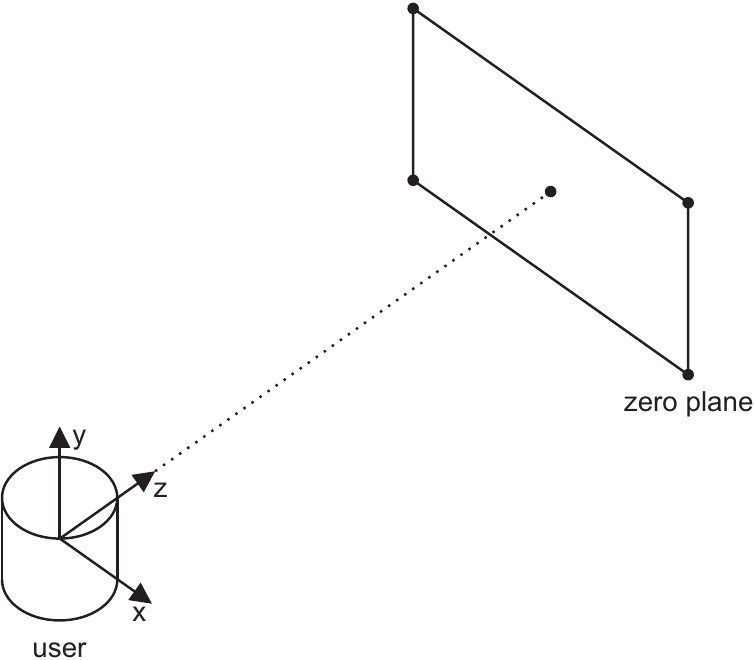}
\caption{Defined coordinate system}
\label{fig:coordinateSystem}
\end{center}
\end{figure}

Let's assume that the size of the computer screen is 21 inch and the aspect ratio is 4:3 (standard display - $43cm \times 32cm$). Having these dimensions we can go step further and investigate what could be a natural head movement in this scenario. Let's define a coordinate system which origin coincides with the origin of the cylinder (the origin of the user's head). Furthermore, let's define that the $x$ axis points towards the rightmost point of the computer screen, the $y$ axis points towards the topmost point of the screen, and the $z$ axis points towards the screen itself. The defined coordinate system is illustrated in figure \ref{fig:coordinateSystem}.

\begin{figure}[htp!]
\begin{center}
\includegraphics[scale=1]{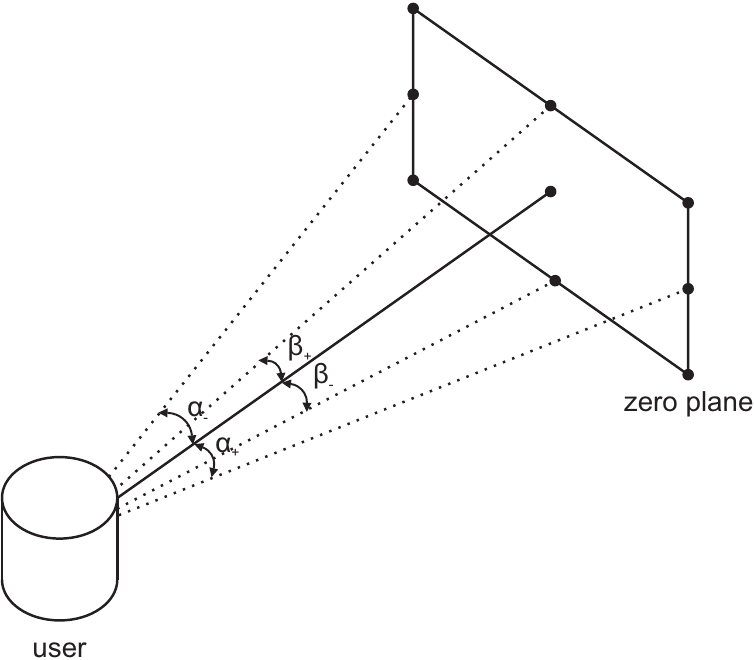}
\caption{Natural head rotation in front of the computer screen}
\label{fig:naturalHeadRotation}
\end{center}
\end{figure}

Having defined the coordinate system, let's have a look at what types of rotation we can observe in the user's behavior. It is highly probable that we are not going to witness huge rotation around the $z$-axis (roll) if we consider natural head movement in front of the computer screen. Most probably, we are going to observe rotation around the $x$ axis (pitch) and the $y$ axis (yaw) where the second one will be bigger (assuming that the width of the screen is bigger than the height). In such constructed 3D space we can calculate what we can call natural head rotation in front of the computer screen. Let's assume that the origin of the cylinder can be projected to the center of the screen and find out what are the angles of these rotations. Figure \ref{fig:naturalHeadRotation} shows this scenario.

\begin{figure}[htp!]
\begin{center}
\subfigure[pitch]{
\includegraphics[scale=1]{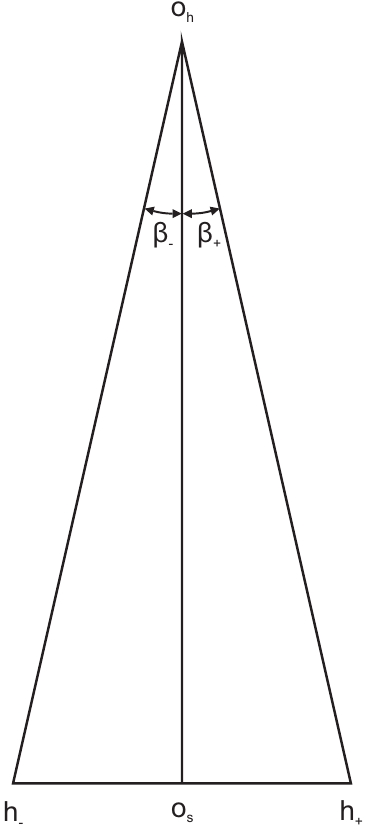}}
\subfigure[yaw]{
\includegraphics[scale=1]{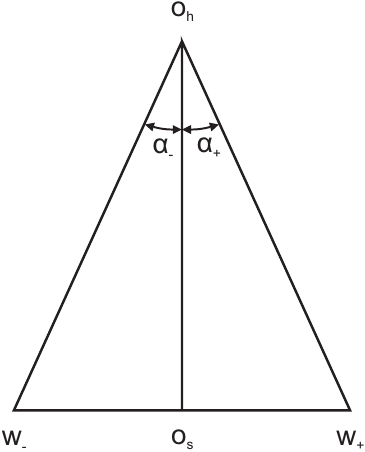}}
\caption{Pitch and yaw}
\label{fig:pitchAndYaw}
\end{center}
\end{figure}

This is a simple mathematical problem and all the steps are shown next. Consider the two problems we need to resolve (pitch and yaw) - their visual representation is shown in figure \ref{fig:pitchAndYaw}, where, $O_{h}$ is the origin of the user's head (cylinder's origin), $O_{s}$ is the center of the computer screen, and $w_{-}$, $w_{+}$, $h_{-}$, $h_{+}$ are the leftmost, rightmost, bottommost and topmost points, respectively, on the screen. Then,

\begin{equation}
\begin{array}{l}
tan(\alpha) = \frac{O_{s}w_{+}}{O_{s}O_{h}} = \frac{215}{750} = 0.2866 \\
\alpha \approx 16.5^{\circ}
\end{array}
\end{equation}

\begin{equation}
\begin{array}{l}
tan(\beta) = \frac{O_{s}h_{+}}{O_{s}O_{h}} = \frac{160}{750} = 0.2133 \\
\beta \approx 12^{\circ}
\end{array}
\end{equation}

These angles give us a glance at what could be considered as natural head rotation in front of the computer screen when the user's head is at distance of $750mm$.
\newpage

Since people tend to position their head in place where the head's origin can be projected close to the center of the screen they do not perform significant amount of head translation (shift), but rotation instead. One can argue with this assumption but consider the following psychological argument - it is reasonable to believe that the user will place his/her head in such a position in order to achieve maximal comfort when using the computer (limitations in the human field of view). Placing the head at this position will ensure that the central $10^{\circ}$ of the visual field will cover as big as possible portion of the computer screen. Furthermore, imagine one sitting in front of the screen and looking at the top-left corner. It is highly probable that one will rotate his/her head towards that point, instead of moving his/her head up and left. This observation shows that it is reasonable to believe that the rotational angles calculated are an approximation on what kind of overall movement we can observe in front of the screen and these angles are the maximal ones so that the center of the head is projected to the point of interest, resulting in projection of the point of interest as close as possible to the central $2^{\circ}$ of both eyes simultaneously (the high accuracy visual perception).

Then, the conducted experiment which we call dynamic do not match these expectations. During the experiment, the test subjects tend to move their head much more (roughly $45^{\circ}$ in $x$ direction and $30^{\circ}$ in $y$ direction) than these rotational angles which results in extreme head movement given the scenario. 

\section{Natural Head Movement}\label{sec:experimentNHM}

These observations and reasoning result in conducting the dynamic experiment in a completely different way. For the new dynamic experiment we are using the animation used for the static experiment (see figure \ref{fig:experimentStatic}). The reason for this is obvious - the point moves around the computer screen and it reaches the most extreme parts of it as well as there is a lot of movement in the inner part of the screen. The test subject is again asked to follow the point, but this time he/she is not restricted in head movement. Actually, the test subject is requested to follow the point's animation around the screen with his/her eyes and head simultaneously. Then, at each time step (25 frames per second) an image of the test subject is recorded while he/she is gazing at the point. The data for test subject is a set of pictures of him/her gazing at different locations on the computer screen while moving his/her head, and the coordinates of the points corresponding to these locations.

The settings of the experiment are the same as before: the test subject's head is at distance of roughly $750mm$ from the computer screen, the camera coordinate system is defined in the middle of the streamed pictures/video and the origin of the subject's head is as close as possible to the origin of the camera coordinate system. The resolution of the captured images is lower than the previous experiments - 720 $\times$ 480 pixels, and the resolution of the computer screen is again 1280 $\times$ 1024 pixels. The moving point is a circle with radius of 25 pixels - resulting in moving a blob on the computer screen that has dimensions of approximately 50 $\times$ 50 pixels. This experiment takes a step further in conforming with the usability requirements because the test subject is never asked to perform calibration at the beginning of the experiment - the experiment is calibration free.

In the video footages we observed that conducting the experiment in this way is justified because the test subject moves his/her head in a natural way and when asked, the experiment did not cause any discomfort, neither for the eyes nor for the head. In fact, it is a simple task of looking around on the computer screen as one would do while sitting in front his/her personal computer and browse the Internet, for example. Furthermore, we observe that a natural head movement in this scenario is indeed bounded by the degrees of rotation discussed earlier and there is no significant amount of head translation.
\cleardoublepage

\chapter{Results}\label{chapter:results}

\dropping{2}{I}\newline n this chapter we present and analyze the results of the conducted experiments. Generally, there are several sources for errors in the visual eye-gaze trackers, so we start with defining the problem again. Then we investigate why the proposed approach introduces errors in the estimate of the point of visual gaze. We continue with the performance of each of the proposed trackers on the data gathered during the described experiments (see chapter \ref{chapter:experiments}). The chapter ends with the conclusions drawn based on the accuracy of the proposed systems.

As stated at the beginning of the manuscript, in 3D space the point of the visual gaze is defined as the intersection of the visual axis of the eye with the computer screen (see figure \ref{fig:gazeDefinitionIntro}). This simple definition proved to be useless in this case because of the restriction on the tracker to be based on a single camera and no prior knowledge for the 3D set-up of the scene. Since we cannot retrieve the depth of the user's head based on the image data of a single camera, an accurate 3D model of the eye cannot be constructed, nor the scene can be accurately modeled. Therefore, an intersection point of the visual axis with the computer screen cannot be defined and computed.

The visual eye-gaze estimators have inherent errors which may occur in each of the components of the visual gaze pipeline (see figure \ref{fig:gazePipeline}). There are two errors (one for each of the components of the pipeline) that can be identified, and which should be taken into account when analyzing the accuracy of proposed trackers: the device error $\epsilon_{d}$ and the calibration error $\epsilon_{c}$. Depending on the scenario, the actual size of the error ($\epsilon_{total}$) is accumulated by the individual contribution of these two errors and the mapping of that error to the distance of the gazed scene.

\begin{itemize}
\item{\textbf{the device error $\epsilon_{d}$} is attributed to the first component of the visual gaze estimation pipeline. Since imaging devices are limited in resolution, there are discrete number of states in which image features can be detected. The variable defining this error is the maximum level of detail which the device can achieve (device resolution) while interpreting pixels as the location of the eyes and/or the position of the head. Therefore, this error depends on the scenario (the distance of the user from the imaging device) and on the device that is being used. Generally, small errors in camera coordinate system result in big errors in the screen coordinate system (further elaborated and quantified in section \ref{sec:resultsConclusion});}
\item{\textbf{the calibration error $\epsilon_{c}$} is attributed to the second component of the visual gaze estimation pipeline. Since eye-gaze trackers use a mapping between the facial feature (head origin)-pupil center vectors and the corresponding location on the computer screen, the tracking system is calibrated at the beginning. In case the user moves from his/her original position during the calibration step, this mapping will be inconsistent and the system may erroneously estimate the point of visual gaze. Chin-rests are often required in this situations to limit the movements of the user to a minimum. Muscular distress, the length of the session, the tiredness of the user, all influence the calibration error. As the calibration error cannot be known a priori, it cannot be modeled. Therefore, some clever method to cope with it is needed (i.e., estimate it, so it can be compensated).}
\end{itemize}

There is a third source of errors in the estimation of the point of visual gaze - the foveal error $\epsilon_{f}$. The fovea is the part of the retina responsible for accurate central vision in the direction in which it is pointed. It is necessary to perform any activities which require a high level of visual details. The human fovea has a diameter of about $1.00mm$ with a high concentration of cone photoreceptors which account for the high visual acuity capability. Through saccades (more than 10000 per hour \cite{GEISLER&BANKS}), the fovea is moved to the regions of interest, generating eye fixations. In fact, if the gazed object is large, the eyes constantly shift their gaze to subsequently bring images into the fovea. For this reason, fixations obtained by analyzing the location of the center of the cornea are widely used in the literature as an indication of the visual gaze and the interest of the user.

However, it is generally assumed that the fixation obtained by analyzing the center of the cornea corresponds to the exact location of interest. While this is valid assumption in most scenarios, the size of the fovea actually permits to see the central two degrees of the visual field. For instance, when reading a text, humans do not fixate on each of the letters, but one fixation permits to read and see multiple words at once.

Another important aspect to be taken into account is the decrease in visual resolution as we move away from the center of the fovea (see figure \ref{fig:acuityHumanEye}). The fovea is surrounded by the parafovea belt which extends up to $1.25mm$ away from the center, followed by the perifovea ($2.75mm$ away), which in turn is surrounded by a larger area that delivers low resolution information. Starting from the outskirts of the fovea, the density of receptors progressively decreases, hence the visual resolution decreases rapidly as it goes far away from the fovea center \cite{ROSSI&ROORDA}.

This comes to support of our previous reasoning that even human is not capable of saying precisely in pixel resolution the location of the point of interest. Therefore, even having minimal device error $\epsilon_{d}$ and calibration error $\epsilon_{c}$, the accuracy of the visual gaze estimation will be bounded by the inherited foveal error $\epsilon_{f}$ stemming from the physiological structure of the human eye.

During the course of this work three main visual eye-gaze trackers were implemented. They are regarded as 2D, 2.5D and 3D in the manuscript, where,

\begin{itemize}
\item{\textbf{2D} is the tracker that uses no information for the position of the user's head (we refer the reader to chapter \ref{chapter:SH} for details on the tracker). The tracker takes into account the location of the eyes and the location of the facial features in camera coordinate system. During the calibration step it constructs vectors between the location of the anchor points (facial features) and the location of the eyes. These vectors are used together with the coordinates of the corresponding points on the computer screen for training. Then, the tracker approximates the coefficients of the underlying model by minimizing an error measure of the misfit of the generated estimates by a candidate model and the train data. When a certain threshold is reached the model is accepted and used for the estimation of the point of visual gaze when a future facial feature-pupil center vector is constructed;}
\item{\textbf{2.5D} is the tracker that uses information for the position of the user's head in the sense that the location of the origin of the cylindrical head model is used as facial features and the location of the eyes is in pose normalized coordinates (refer to chapter \ref{chapter:NHM} for details on the tracker). This tracker should perform better than the 2D one because the anchor points (cylinder's origin) are more stable than the manually defined one in the 2D tracker, resulting in more stable cylinder origin-pupil center vectors. The rest of the system is the same as in the 2D tracker;}
\item{\textbf{3D} is the tracker that is proposed by this manuscript for estimation of the point of visual gaze under natural head movement. The tracker uses the location of the origin of the cylindrical head model as facial features and the location of the eyes are in pose normalized coordinates (same as the 2.5D tracker). As it was described, this tracker implements the idea that each movement of the user's head can be treated as a special two-dimensional case, so the set of calibration points is updated each time the user moves his/her head (we refer the reader to chapter \ref{chapter:NHM} for details on the system). The main difference between the 3D tracker and the 2.5D one is that when the user moves his/her head the set of calibration points is updated and the system is re-trained. Then new coefficients of the underlying model are approximated and used for the estimate of the point of visual gaze.}
\end{itemize}
\newpage

There are two main classes of calibration methods presented in the manuscript; one that uses 2 calibration points and a linear model and one that uses 5 (linear model) and 9 or 25 (second order model) calibration points. Since the solution in the second method is achieved through a least squares approximation it is reasonable to believe that the amount of data points will govern the accuracy of the approximation. This is the main reason to omit the calibration with 9 points, and use 25 points in this method. Furthermore, it is reasonable to believe that a second order model is capable to capture the function behavior better than a linear model; then, we discard the calibration with 5 points as well in the testing phase.

We conducted three experiments to test our ideas. The first experiment is testing the accuracy of the proposed trackers when the test subject is requested to keep his/her head as static as possible (refer to section \ref{sec:static} for details on the experiment's design). In the second experiment the test subject is requested to move his/her head in specific way (refer to section \ref{sec:dynamic}). As reasoned in the previous chapter this experiment showed that the test subjects are performing extreme head movements given the scenario. Then, the last experiment performed is based on the observations made during the extreme head movement experiment (refer to section \ref{sec:experimentNHM} for detailed description of the experiment's design). The rest of the chapter is structured in the same order where the performance of each of the trackers is presented for each of the experiments. A comparison between the different trackers and calibration methods is offered for each of the experiments as well. It is a fair comparison because the same data is used for each of the compared trackers/calibration methods.

\section{No Head Movement}\label{sec:SH}

All results in this section are based on the data gathered during the experiment in which the test subjects are requested to keep their head as static as possible (refer to section \ref{sec:static}) and follow an animation of a point on the computer screen only with their eyes. The two tables below present detailed overview of the error of the visual gaze estimation for each test subject for each of the three trackers. The trackers are compared by computing the mean error they have - the lower the mean error the better the tracker. Furthermore, the tables present the results for the two calibration methods, so we can investigate how the choice of calibration method influences the accuracy of each of the proposed systems.
\newpage

\begin{table}[htbp!]\tiny
\begin{center}
\begin{tabular}{|c|c|c|c|c|c|c|}
\hline
\multicolumn{7}{|c|}{\textbf{2D vs 2.5D vs 3D}} \\ \hline
\textbf{Subject} & \multicolumn{2}{|c|}{\textbf{2D}} & \multicolumn{2}{|c|}{\textbf{2.5D}} & \multicolumn{2}{|c|}{\textbf{3D}} \\ \hline
\textbf{} & \textbf{Mean} & \textbf{Std} & \textbf{Mean} & \textbf{Std} & \textbf{Mean} & \textbf{Std} \\ \hline
1 & 271.91, 73.74 & 120.39, 54.93 & 128.29, 82.42 & 61.93, 57.59 & 90.72, 165.98 & 52.69, 88.66 \\ \hline
2 & 272.13, 74.71 & 174.92, 60.23 & 104.76, 74.68 & 62.57, 57.42 & 78.93, 110.95 & 53.01, 72.12 \\ \hline
3 & 269.11, 96.05 & 643.46, 198.38 & 621.51, 226.25 & 357.53, 103.18 & 621.03, 311.18 & 344.64, 91.91 \\ \hline
4 & 448.56, 317.26 & 775.57, 1134.81 & 171.02, 127.61 & 82.94, 79.97 & 126.67, 175.23 & 84.47, 91.81 \\ \hline
5 & 453.31, 98.76 & 313.59, 62.72 & 847.58, 105.51 & 372.41, 65.22 & 851.03, 127.28 & 405.94, 71.71 \\ \hline
6 & 315.27, 97.65 & 166.49, 67.03 & 238.81, 88.71 & 81.87, 69.14 & 199.57, 84.55 & 78.85, 65.08 \\ \hline
7 & 163.26, 124.66 & 148.24, 69.82 & 423.71, 86.75 & 273.51, 63.16 & 483.87, 98.61 & 262.28, 68.97 \\ \hline
8 & 170.48, 82.16 & 135.11, 55.27 & 500.16, 291.91 & 327.32, 125.91 & 557.58, 263.84 & 306.18, 106.34 \\ \hline
9 & 174.44, 124.41 & 158.03, 90.47 & 474.84, 309.11 & 293.31, 197.14 & 471.45, 348.61 & 273.68, 197.46 \\ \hline
10 & 189.03, 93.93 & 254.06, 347.13 & 204.03, 68.67 & 110.11, 44.11 & 211.56, 101.18 & 129.56, 55.63 \\ \hline
11 & 242.43, 58.49 & 178.81, 60.65 & 601.07, 651.39 & 358.44, 291.53 & 650.83, 643.25 & 326.46, 275.01 \\ \hline
\end{tabular}
\end{center}
\caption{Trackers' accuracy comparison using 2 calibration points under no head movement}
\label{table:comparisonStatic2P}
\end{table}

\begin{table}[htbp!]\tiny
\begin{center}
\begin{tabular}{|c|c|c|c|c|c|c|}
\hline
\multicolumn{7}{|c|}{\textbf{2D vs 2.5D vs 3D}} \\ \hline
\textbf{Subject} & \multicolumn{2}{|c|}{\textbf{2D}} & \multicolumn{2}{|c|}{\textbf{2.5D}} & \multicolumn{2}{|c|}{\textbf{3D}} \\ \hline
\textbf{} & \textbf{Mean} & \textbf{Std} & \textbf{Mean} & \textbf{Std} & \textbf{Mean} & \textbf{Std} \\ \hline
1 & 103.51, 69.38 & 88.53, 58.23 & 47.75, 69.71 & 36.29, 51.11 & 62.39, 181.62 & 41.75, 75.49 \\ \hline
2 & 106.91, 80.41 & 94.66, 65.18 & 47.34, 54.83 & 36.93, 49.82 & 42.96, 82.01 & 33.98, 65.61 \\ \hline
3 & 164.55, 98.62 & 312.31, 268.13 & 62.46, 49.91 & 46.26, 44.62 & 57.48, 97.21 & 46.82, 64.51 \\ \hline
4 & 370.09, 697.86 & 1725.01, 4483.81 & 62.67, 134.35 & 48.86, 112.67 & 72.03, 105.74 & 54.04, 71.19 \\ \hline
5 & 151.32, 79.15 & 139.86, 51.68 & 66.82, 66.67 & 52.12, 52.38 & 49.39, 75.75 & 37.46, 58.67 \\ \hline
6 & 100.48, 85.04 & 89.69, 56.47 & 45.31, 69.11 & 34.42, 55.59 & 52.05, 106.05 & 37.25, 78.31 \\ \hline
7 & 108.83, 68.31 & 95.26, 51.91 & 65.87, 77.15 & 46.96, 68.04 & 59.31, 86.29 & 44.35, 67.49 \\ \hline
8 & 123.92, 112.07 & 116.08, 73.09 & 63.49, 75.42 & 49.11, 50.11 & 74.05, 87.08 & 61.14, 58.25 \\ \hline
9 & 109.01, 84.35 & 114.49, 60.35 & 49.81, 70.54 & 36.96, 50.07 & 50.45, 102.29 & 37.16, 69.91 \\ \hline
10 & 125.64, 73.74 & 207.76, 159.42 & 58.74, 58.54 & 41.57, 44.21 & 61.45, 68.95 & 43.97, 49.77 \\ \hline
11 & 126.46, 60.25 & 99.39, 54.97 & 56.21, 52.84 & 38.35, 46.15 & 59.21, 56.08 & 46.77, 51.91 \\ \hline
\end{tabular}
\end{center}
\caption{Trackers' accuracy comparison using 25 calibration points under no head movement}
\label{table:comparisonStatic25P}
\end{table}

Table \ref{table:comparisonStatic2P} presents the trackers comparison based on the calibration method that uses 2 calibration points and a linear model. The 2D tracker has a mean error of (269.99, 112.89) pixels in $x$ and $y$ direction, respectively, the 2.5D tracker has a mean error of (392.34, 192.09) pixels in $x$ and $y$ direction, respectively, and the 3D tracker has a mean error of (394.84, 220.96) pixels in $x$ and $y$ direction, respectively. The result shows that when using this calibration method, the 2D tracker achieves highest accuracy. Although, the 2D tracker is designed for this settings (no head movement), the result is unexpected. The design of the 2.5D tracker implies that it should perform at least as good as or better than the 2D one. Since, as we can see in the next results (using 25 calibration points in the same settings), we cannot say that the device error is the reason for the fact that the 2D tracker performs better than the 2.5D one, the only explanation is the behavior of the linear model given the facial feature-pupil center vectors. The 3D tracker is intended to cope with large head movements, so we can expect that its performance will be as good as the 2.5D one or worse in this settings (because the re-calibration step introduces errors in the estimate).

\begin{figure}[htp!]
\begin{center}
\includegraphics[scale=1]{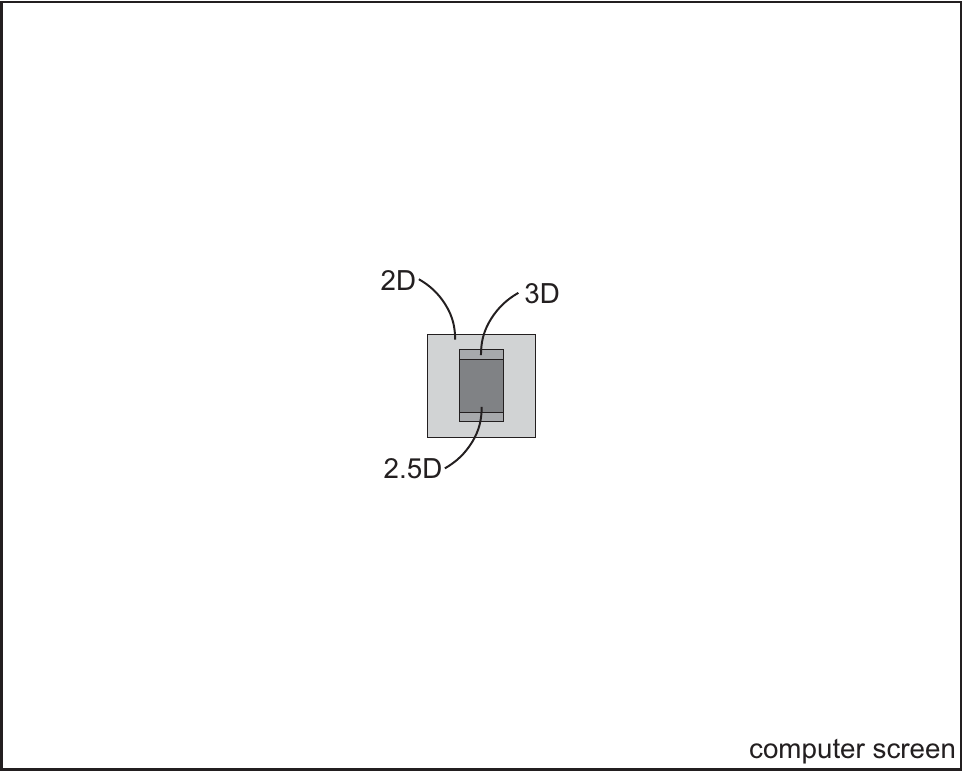}
\caption{Errors compared to the computer screen (no head movement)}
\label{fig:staticError}
\end{center}
\end{figure}

Table \ref{table:comparisonStatic25P} compares the three algorithm based on the calibration method that uses 25 calibration points and a second order model. In this settings the 2D tracker commits a mean error of (144.61, 137.19), the 2.5D tracker commits a mean error of (56.95, 70.82), and the 3D tracker commits a mean error of (58.25, 95.37), all in pixels, in $x$ and $y$ direction, respectively. This result shows that when using the calibration method with 25 points, the 2.5D tracker performs best. It is an expected result because as mentioned previously this tracker is designed for this scenario. The result shows that the more stable facial feature-pupil center vectors of the 2.5D tracker increase the accuracy with a factor of about 2.53 in $x$ direction and with a factor of about 1.93 in $y$ direction compared to the entirely 2D approach. It is interesting to see that the 3D tracker performs good (close to the 2.5D one), although its main purpose is to cope with large head movements. Figure \ref{fig:staticError} gives the visual representation of the mean error window of each tracker in this settings (no head movement and calibration with 25 points and a second order model) compared to the computer screen.

Finally, the results show that the calibration with 25 points outperforms the one with 2 points in this settings (no head movement). Although, the error of the 2D tracker in $y$ direction is higher when using 25 calibration points, close investigation of the tables shows that there is only one case (test subject 4 in table \ref{table:comparisonStatic25P}) with high error. Since, the only difference is the calibration method, this is a good example of how a bad calibration procedure (the more complicated one with 25 points) introduces big calibration error and influences the accuracy of the estimate for the point of visual gaze.

\section{Extreme Head Movement}\label{sec:EHM}

In this section we compare the results of the three trackers based on the data gathered during the dynamic experiment (refer to section \ref{sec:dynamic} for the design of the experiment); the test subjects are requested to look at a static point on the computer screen and say `yes' and `no' with their head. As reasoned previously (see section \ref{sec:experimentsConclusions}), in this experiment the test subjects perform what can be regarded as an extreme head movement given the scenario. The two tables below present the error that each of the trackers commits. The trackers are again compared by computing the mean error they have for both calibration methods.

\begin{table}[htbp!]\tiny
\begin{center}
\begin{tabular}{|c|c|c|c|c|c|c|}
\hline
\multicolumn{7}{|c|}{\textbf{2D vs 2.5D vs 3D}} \\ \hline
\textbf{Subject} & \multicolumn{2}{|c|}{\textbf{2D}} & \multicolumn{2}{|c|}{\textbf{2.5D}} & \multicolumn{2}{|c|}{\textbf{3D}} \\ \hline
\textbf{} & \textbf{Mean} & \textbf{Std} & \textbf{Mean} & \textbf{Std} & \textbf{Mean} & \textbf{Std} \\ \hline
1 & 636.51, 335.92 & 458.05, 410.72 & 350.38, 314.66 & 169.89, 210.71 & 178.86, 175.18 & 195.45, 148.11 \\ \hline
2 & 674.81, 164.18 & 483.05, 142.16 & 327.13, 218.02 & 219.27, 174.73 & 173.36, 194.73 & 189.16, 142.18 \\ \hline
3 & 2227.33, 590.95 & 4918.12, 1079.51 & 479.55, 468.34 & 412.73, 281.55 & 468.56, 482.34 & 297.08, 207.33 \\ \hline
4 & 986.62, 824.92 & 1474.53, 1737.91 & 324.51, 277.44 & 178.71, 210.27 & 188.61, 167.71 & 192.31, 133.46 \\ \hline
5 & 945.29, 208.74 & 713.39, 376.67 & 875.23, 304.89 & 483.12, 222.61 & 699.59, 195.46 & 445.35, 170.75 \\ \hline
6 & 665.61, 232.86 & 468.55, 366.36 & 461.98, 269.71 & 276.23, 187.19 & 257.11, 200.81 & 209.78, 137.04 \\ \hline
7 & 660.83, 239.21 & 362.87, 184.25 & 519.06, 236.97 & 420.72, 185.22 & 604.11, 173.52 & 296.58, 113.85 \\ \hline
8 & 583.52, 357.65 & 312.55, 245.24 & 458.31, 256.74 & 425.81, 224.43 & 524.19, 245.41 & 308.77, 170.46 \\ \hline
9 & 625.89, 360.22 & 408.87, 216.06 & 463.65, 375.31 & 360.06, 330.94 & 377.56, 304.88 & 261.53, 299.25 \\ \hline
10 & 587.73, 214.97 & 410.06, 487.02 & 440.14, 270.42 & 295.54, 218.76 & 277.06, 203.33 & 252.39, 171.86 \\ \hline
11 & 728.69, 468.99 & 579.65, 628.01 & 490.85, 837.33 & 397.51, 577.42 & 447.82, 571.31 & 299.31, 442.53 \\ \hline
\end{tabular}
\end{center}
\caption{Trackers' accuracy comparison using 2 calibration points under extreme head movement}
\label{table:comparisonDynamic2P}
\end{table}

\begin{table}[htbp!]\tiny
\begin{center}
\begin{tabular}{|c|c|c|c|c|c|c|}
\hline
\multicolumn{7}{|c|}{\textbf{2D vs 2.5D vs 3D}} \\ \hline
\textbf{Subject} & \multicolumn{2}{|c|}{\textbf{2D}} & \multicolumn{2}{|c|}{\textbf{2.5D}} & \multicolumn{2}{|c|}{\textbf{3D}} \\ \hline
\textbf{} & \textbf{Mean} & \textbf{Std} & \textbf{Mean} & \textbf{Std} & \textbf{Mean} & \textbf{Std} \\ \hline
1 & 688.91, 281.29 & 1090.71, 552.12 & 338.97, 366.42 & 181.04, 280.57 & 186.11, 223.51 & 191.32, 212.11 \\ \hline
2 & 633.08, 187.74 & 464.23, 143.99 & 310.17, 226.07 & 182.89, 201.83 & 134.91, 191.77 & 197.01, 153.41 \\ \hline
3 & 2285.11, 301.45 & 4929.71, 436.26 & 359.97, 246.45 & 184.91, 216.05 & 161.21, 255.71 & 201.04, 168.91 \\ \hline
4 & 1202.11, 2664.01 & 2537.21, 7280.43 & 346.56, 330.79 & 164.11, 246.59 & 190.12, 164.84 & 199.37, 154.58 \\ \hline
5 & 1388.72, 276.79 & 1073.91, 630.76 & 428.04, 327.46 & 222.63, 287.75 & 239.25, 215.67 & 200.78, 225.44 \\ \hline
6 & 874.85, 239.81 & 726.24, 491.89 & 429.17, 265.61 & 215.31, 190.13 & 232.84, 234.02 & 175.45, 177.93 \\ \hline
7 & 710.24, 328.63 & 443.08, 224.25 & 449.44, 217.21 & 240.39, 175.33 & 242.87, 162.01 & 188.21, 140.37 \\ \hline
8 & 666.58, 257.17 & 376.31, 194.65 & 397.56, 236.25 & 226.65, 181.82 & 196.38, 152.45 & 191.22, 131.81 \\ \hline
9 & 623.68, 316.73 & 412.11, 209.09 & 395.59, 337.47 & 206.61, 246.04 & 204.87, 220.94 & 197.46, 202.01 \\ \hline
10 & 750.93, 332.06 & 947.91, 1462.83 & 430.95, 319.41 & 247.44, 263.63 & 272.23, 223.17 & 231.42, 205.61 \\ \hline
11 & 924.62, 398.24 & 2297.01, 297.93 & 443.03, 580.46 & 229.99, 412.91 & 252.93, 320.81 & 186.16, 255.22 \\ \hline
\end{tabular}
\end{center}
\caption{Trackers' accuracy comparison using 25 calibration points under extreme head movement}
\label{table:comparisonDynamic25P}
\end{table}

Table \ref{table:comparisonDynamic2P} presents a performance comparison based on the calibration method that uses 2 calibration points and a linear model, in pixels, in $x$ and $y$ direction, respectively. The 2D system has a mean error of (847.53, 363.51), the 2.5D system has a mean error of (471.89, 348.16), and the 3D system has a mean error of (381.53, 264.97). These results show the potential of the proposed 3D tracker. The 3D tracker achieves an improvement with a factor of about 1.23 in $x$ direction and improvement with a factor of about 1.31 in $y$ direction compared to the 2.5D system. As expected, the 2D tracker fails to achieve any meaningful results because the position of the test subject's head differs from the position at which the calibration is performed. The 2.5D tracker improves the accuracy of the estimate with a factor of about 1.79 in $x$ direction and with factor of about 1.04 in $y$ direction compared to the 2D one. This improvement is attributed to the fact the 2.5D tracker uses the origin of the cylindrical head model as facial features, resulting in more stable vectors; furthermore, in many frames the 2D tracker loses the facial features location because of the extreme head movement.

\begin{figure}[htp!]
\begin{center}
\includegraphics[scale=1]{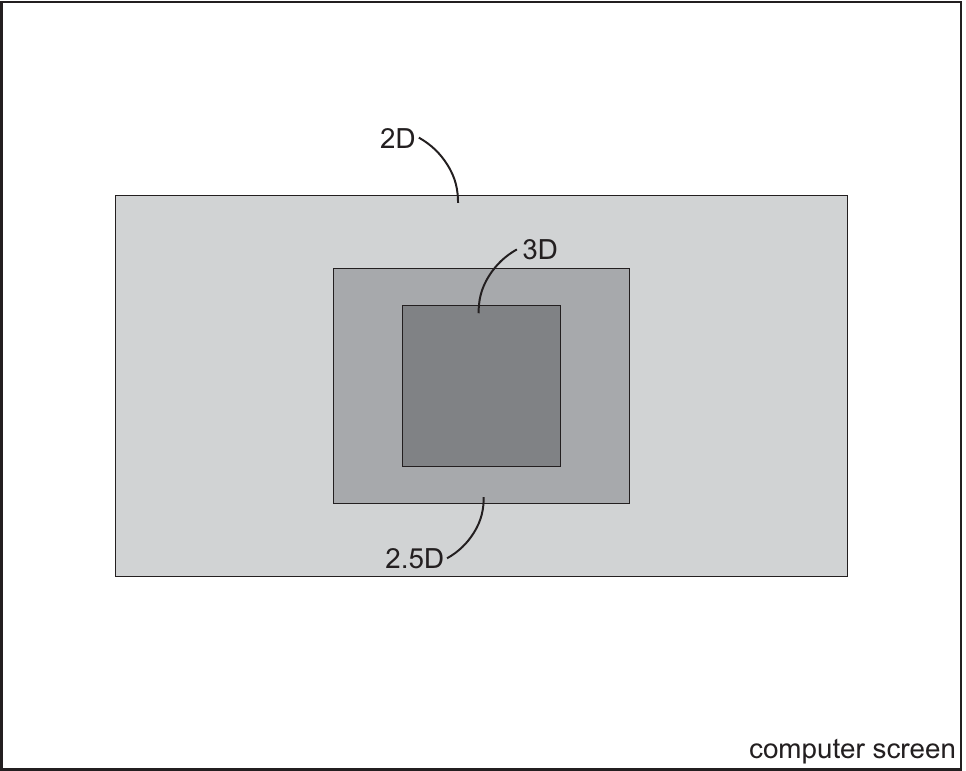}
\caption{Errors compared to the computer screen (extreme head movement)}
\label{fig:unnaturalHMError}
\end{center}
\end{figure}

In table \ref{table:comparisonDynamic25P} we examine the performance of the trackers when a calibration method with 25 calibration points and a second order model is used. In this settings, the 2D tracker has a mean error of (977.16, 507.62), the 2.5D tracker has a mean error of (393.58, 313.96), and the 3D tracker has a mean error of (210.33, 214.99). All the results are again in pixels, in $x$ and $y$ direction, respectively. The 3D tracker is capable of increasing the accuracy with factor of approximately 1.87 in $x$ direction and to improve with a factor of about 1.46 in $y$ direction compared to the 2.5D system. In essence this improvement is big - consider the result for the 2.5D tracker; it divides the computer screen into 3 $\times$ 3 grid on average, while the 3D tracker divides the screen into 6 $\times$ 5 grid. Furthermore, the results of the 3D tracker are more stable (standard deviation) then the one of the 2.5D tracker. The 2D system fails to achieve any meaningful results for the same reasons as in the scenario with 2 calibration points. For these reasons the 2.5D tracker improves even more the accuracy of the estimate compared to the 2D system; with a factor of about 2.48 in $x$ direction and with a factor of about 1.61 in $y$ direction. The mean error window of each tracker in this settings (25 calibration points and extreme head movement) is shown in figure \ref{fig:unnaturalHMError}.

These results again show that the calibration with 25 points outperforms the one with 2 points in both 2.5D and 3D trackers. This is not valid for the 2D tracker anymore and in this settings we see that this result is a definite one (contrary to the one observed in the static head scenario). We can conclude that the errors committed by the linear model in these extreme conditions (tracker wise) are less than these committed by the second order model. Nevertheless, this observation has no impact because the 2D system still performs extremely poor.

\section{Natural Head Movement}\label{sec:NHM}

The comparisons in this section are based on the data generated during the last experiment conducted. As described in section \ref{sec:experimentsConclusions} some observations were made during the dynamic experiment which inspired a new design for it (refer to section \ref{sec:experimentNHM} for details); the test subjects are requested to look at animation on the computer screen and follow a moving point with their eyes and head simultaneously. The errors of each tracker in this settings is presented in the two tables below.

\begin{table}[htbp!]\tiny
\begin{center}
\begin{tabular}{|c|c|c|c|c|c|c|}
\hline
\multicolumn{7}{|c|}{\textbf{2D vs 2.5D vs 3D}} \\ \hline
\textbf{Subject} & \multicolumn{2}{|c|}{\textbf{2D}} & \multicolumn{2}{|c|}{\textbf{2.5D}} & \multicolumn{2}{|c|}{\textbf{3D}} \\ \hline
\textbf{} & \textbf{Mean} & \textbf{Std} & \textbf{Mean} & \textbf{Std} & \textbf{Mean} & \textbf{Std} \\ \hline
1 & 1485.21, 360.41 & 969.35, 226.91 & 277.96, 142.11 & 183.86, 95.31 & 216.13, 75.37 & 102.44, 49.07 \\ \hline
2 & 1448.22, 496.99 & 2762.61, 896.57 & 255.88, 121.53 & 187.85, 91.31 & 202.98, 75.85 & 86.64, 53.21 \\ \hline
3 & 1911.49, 412.39 & 3751.34, 976.13 & 277.28, 103.11 & 214.91, 73.48 & 227.99, 69.61 & 88.72, 54.46 \\ \hline
4 & 1594.86, 422.23 & 1204.31, 434.19 & 308.21, 140.45 & 275.05, 96.91 & 247.47, 85.41 & 130.01, 100.11 \\ \hline
5 & 1435.91, 329.55 & 959.59, 247.07 & 297.49, 130.01 & 227.37, 102.01 & 228.55, 63.54 & 97.01, 46.81 \\ \hline
6 & 1502.01, 343.19 & 1031.67, 313.21 & 309.82, 145.88 & 214.62, 82.78 & 255.26, 60.56 & 106.99, 45.97 \\ \hline
7 & 1834.57, 473.37 & 1543.83, 804.96 & 333.58, 125.07 & 247.67, 85.35 & 257.13, 67.52 & 112.77, 55.22 \\ \hline
8 & 1762.06, 536.91 & 3428.56, 359.18 & 318.61, 245.06 & 199.61, 166.53 & 223.06, 115.57 & 107.59, 83.36 \\ \hline
9 & 5191.66, 853.56 & 9109.08, 422.23 & 354.45, 256.38 & 196.12, 175.37 & 291.01, 131.62 & 139.48, 100.54 \\ \hline
10 & 3241.95, 916.71 & 6067.62, 355.55 & 325.93, 224.51 & 237.14, 141.64 & 239.46, 111.36 & 113.57, 79.05 \\ \hline
11 & 5011.67, 503.11 & 8296.25, 326.97 & 380.68, 223.51 & 263.28, 164.91 & 298.28, 117.23 & 120.56, 105.47 \\ \hline
\end{tabular}
\end{center}
\caption{Trackers' accuracy comparison using 2 calibration points under natural head movement}
\label{table:comparisonNewExperiment2P}
\end{table}

\begin{table}[htbp!]\tiny
\begin{center}
\begin{tabular}{|c|c|c|c|c|c|c|}
\hline
\multicolumn{7}{|c|}{\textbf{2D vs 2.5D vs 3D}} \\ \hline
\textbf{Subject} & \multicolumn{2}{|c|}{\textbf{2D}} & \multicolumn{2}{|c|}{\textbf{2.5D}} & \multicolumn{2}{|c|}{\textbf{3D}} \\ \hline
\textbf{} & \textbf{Mean} & \textbf{Std} & \textbf{Mean} & \textbf{Std} & \textbf{Mean} & \textbf{Std} \\ \hline
1 & 3461.68, 938.55 & 1931.83, 567.68 & 238.88, 112.91 & 159.53, 69.42 & 75.95, 117.01 & 71.02, 76.82 \\ \hline
2 & 3125.42, 361.55 & 5874.41, 253.68 & 229.78, 104.72 & 137.44, 73.58 & 79.16, 115.87 & 58.79, 82.01 \\ \hline
3 & 3531.19, 564.11 & 7725.46, 353.82 & 253.51, 103.87 & 162.11, 77.08 & 78.73, 128.01 & 67.48, 77.07 \\ \hline
4 & 2380.21, 400.89 & 2002.35, 608.31 & 277.71, 134.53 & 180.27, 139.32 & 99.29, 115.16 & 108.59, 150.38 \\ \hline
5 & 3554.94, 656.51 & 2799.42, 468.44 & 268.51, 105.09 & 165.54, 77.78 & 85.77, 101.13 & 78.03, 72.79 \\ \hline
6 & 2365.84, 472.37 & 1574.86, 336.32 & 254.63, 95.47 & 165.61, 62.83 & 80.25, 78.27 & 78.67, 57.13 \\ \hline
7 & 3606.85, 729.86 & 3414.06, 1730.97 & 282.74, 101.62 & 179.79, 76.36 & 93.81, 104.64 & 84.21, 81.32 \\ \hline
8 & 3332.96, 573.66 & 6989.07, 625.28 & 278.79, 188.84 & 189.94, 137.01 & 92.25, 92.77 & 85.31, 65.44 \\ \hline
9 & 11958.67, 775.88 & 21508.32, 752.94 & 250.25, 200.03 & 180.95, 139.38 & 92.52, 99.32 & 72.27, 71.75 \\ \hline
10 & 5082.26, 731.92 & 9972.58, 719.33 & 295.22, 168.61 & 190.91, 115.41 & 91.92, 92.21 & 86.09, 59.38 \\ \hline
11 & 8482.27, 693.31 & 13728.11, 832.69 & 303.83, 179.69 & 201.23, 162.51 & 89.36, 98.12 & 109.38, 138.84 \\ \hline
\end{tabular}
\end{center}
\caption{Trackers' accuracy comparison using 25 calibration points under natural head movement}
\label{table:comparisonNewExperiment25P}
\end{table}

In table \ref{table:comparisonNewExperiment2P} the error of the three trackers using the calibration with 2 points and a linear model is presented. The mean error of 2D tracker is (2401.78, 513.49), the one of the 2.5D tracker is (312.71, 168.87), and the mean error of the 3D tracker is (244.31, 88.51) in pixels, in $x$ and $y$ direction, respectively. The results show that the 3D system achieves an improvement with a factor of about 1.27 in $x$ direction and improvement with a factor of about 1.91 in $y$ direction compared to the 2.5D system. The 2D tracker fails even more in this settings for the same reasons as before. The improvement of the 2.5D tracker is with a factor of about 7.68 in $x$ direction and with factor of about 3.04 in $y$ direction compared to the 2D one.

\begin{figure}[htp!]
\begin{center}
\includegraphics[scale=1]{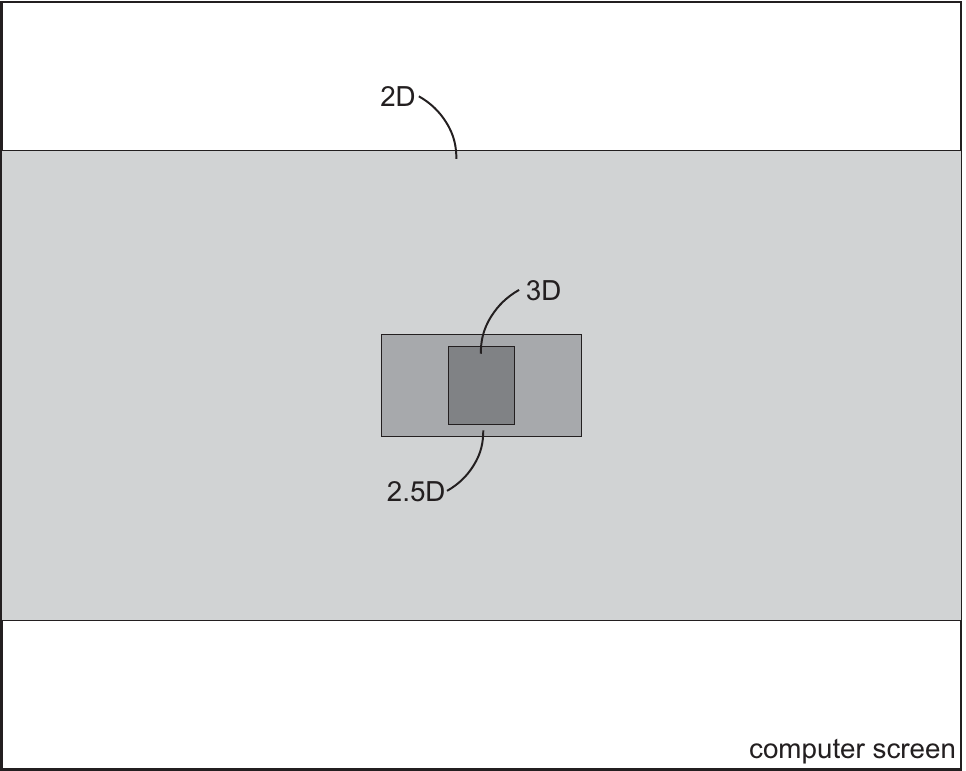}
\caption{Errors compared to the computer screen (natural head movement)}
\label{fig:naturalHMError}
\end{center}
\end{figure}

Table \ref{table:comparisonNewExperiment25P} presents the results when the calibration method with 25 points and a second order model is used. The 2D tracker has a mean error of (4625.66, 627.14), the 2.5D tracker has a mean error of (266.71, 135.94), and the 3D tracker has a mean error of (87.18, 103.86) in pixels, in $x$ and $y$ direction, respectively. In this settings, the 3D system is capable of increasing the accuracy with factor of approximately 3.05 in $x$ direction and with a factor of about 1.31 in $y$ direction compared to the 2.5D system. This improvement is quite big - the result for the 2.5D system divides the computer screen into 5 $\times$ 7.5 grid on average, while the 3D system divides the screen into 15 $\times$ 10 grid. Also the result of the 3D tracker is more stable then the one of the 2.5D system. The 2.5D tracker improves drastically the accuracy compared to the 2D one - with a factor of about 17.34 in $x$ direction and with factor of about 4.61 in $y$ direction. Figure \ref{fig:naturalHMError} gives a visual representation of the mean error window of each tracker in this settings.

These results show that the calibration with 25 points outperforms the one with 2 points in this settings for the 2.5D and the 3D systems. This is again not valid for the 2D tracker anymore. Although, in the experiment in which the test subject's head is static, we see that this fact is attributed to only one subject and his/her erroneous calibration, this result is more strong in settings of dynamic head (extreme and natural movement). We can conclude that the errors committed by the linear model are less than these committed by the second order model when the subject moves his/her head. And again, this observation has no important impact because the 2D system performs even worse than in the previous experiment.

\section{Conclusions}\label{sec:resultsConclusion}

Based on the conducted experiments we can conclude that the calibration method that uses 25 calibration points and a second order model outperforms the method that uses 2 calibration points and a linear model. We observed this behavior in the 2.5D and the 3D trackers in all experiments conducted. We observed that this is valid for the 2D tracker in settings with static head as well. This observation is not true for the 2D tracker in settings with dynamic head (extreme and natural movement) but then the tracker's behavior is unpredictable which results in extremely bad performance.

It is worth mentioning that in most cases the results in $y$ direction are worse than the results in $x$ direction. There are two main reasons for this behavior: the camera is situated on top of the computer screen so when the test subject is gazing at the bottom part of the screen the eyelids obscure the eye location and errors are introduced in the eye tracker (there is only one way to overcome this problem - the camera origin must coincide with the center of the screen); the second problem is the fact that the eye moves less in $y$ direction than in $x$ direction which introduces higher uncertainty. Considering this, we can expect bigger factor of improvement of the 3D system in $x$ direction than in $y$ direction, as we observed.

Investigation on the origins of the errors rises again the question for the device error $\epsilon_{d}$, the calibration error $\epsilon_{c}$ and the the foveal error $\epsilon_{f}$ and their magnitude in the experiments. The device error $\epsilon_{d}$ in these experiments is high; generally, there are two aspects of the device error that should be taken under consideration,

\begin{itemize}
\item{the first aspect is the resolution of the device itself - the static experiment showed that the eye shifts of approximately 10 pixels horizontally and 8 pixels vertically while the test subject is looking at the extremes of the computer screen. Therefore, when looking at a point on the computer screen with resolution of 1280 $\times$ 1024 pixels, there will be an uncertainty window of 128 $\times$ 128 pixels, associated with that point;}
\item{the second aspect is the detection error - this error has two aspects itself - the errors committed by the eye center locator and the errors committed by the head tracker. The used system for eye center location \cite{VALENTI&GEVERS} claims an accuracy close to 100\% for the eye center being located within 5\% of the interocular distance and the head tracker accuracy is discussed in \cite{VALENTI&YUCEL&GEVERS}. These errors will expand even further the uncertainty window associated with the maximum shift of the eye in the images obtained through the low resolution web camera.}
\end{itemize}

The calibration error $\epsilon_{c}$ is hard to quantify in this scenario. It is evident from the results that there is big calibration error in some of the test subjects. Since one of the goals of the proposed eye-gaze tracker is to offer a calibration free set-up the data needed to ensure minimal calibration error should be gathered through strictly controlled experiment (further elaborated in section \ref{sec:improvements}).

\begin{figure}[htp!]
\begin{center}
\includegraphics[scale=1]{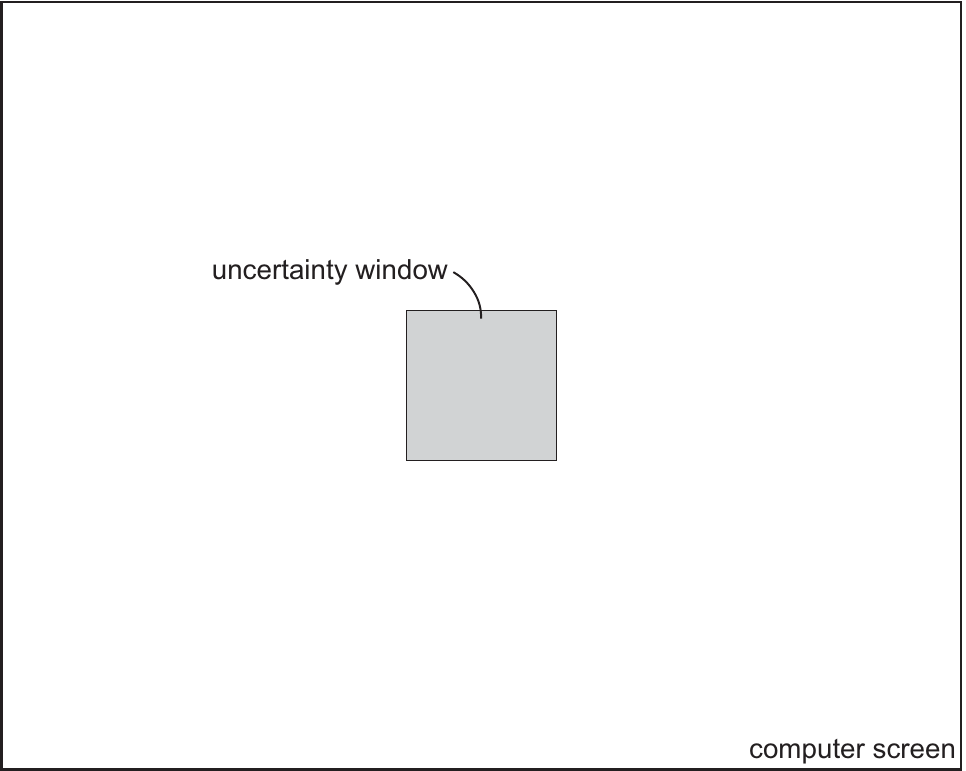}
\caption{Inherited uncertainty window}
\label{fig:uncertaintyWindow}
\end{center}
\end{figure}

Assuming that the test subjects are sitting at distance of $750mm$ from the computer screen, the projection of the $\epsilon_{f} = 2^{\circ}$ corresponds to about 92 $\times$ 92 pixels window. Summing the errors gives a glance in the expected uncertainty window on the screen, or $\epsilon_{total}$ is about 200 $\times$ 200 pixels. Figure \ref{fig:uncertaintyWindow} gives a visual representation on how big the uncertainty windows is compared to the computer screen.

\cleardoublepage

\chapter{Conclusions}\label{chapter:conclusions}

\dropping{2}{I}\newline n this chapter we propose directions for improvements and future work in section \ref{sec:improvements}. The main conclusions are drawn in section \ref{sec:conclusions}. The chapter ends with the main contributions (section \ref{sec:contributions}) of the project.

\section{Improvements and Future Work}\label{sec:improvements}

Although the results show that there is an improvement in the accuracy of the visual gaze estimation stemming from the proposed tracker, there is still a room for improvements. As investigated previously, the total error rate is accumulated by the device error, the calibration error and the foveal error. Efficient ways to cope and reduce the magnitude of these errors, need further investigation.

The device error $\epsilon_{d}$ is affected by several variables:
\begin{itemize}
\item{device resolution - it is evident that the resolution of the device is of huge importance for the device error. We can find many examples in the literature where, researchers use high definition cameras or other specialized equipment in order to minimize the contribution of the device resolution to the total device error. Since this project is concerned with building an accurate system using an ordinary web camera new techniques for coping with this drawback need investigation. One way to go is to use the theory of super-resolution. The authors of \cite{GLASNER&BAGON&IRANI} propose an accurate algorithm for super-resolution from a single image. Enhancement of the image features using similar techniques can increase the search space for the eye center location resulting in more accurate detection;}
\item{eye tracking errors - during the experiments we observed that the used eye tracker achieves good results. Nevertheless, it was observed that the accuracy of the eye tracker depends on the illumination conditions. The algorithm itself is already designed to cope with different illumination conditions through adjustments of some of its parameters, although it is not obvious which parameters are good for which illumination conditions. A step in direction of automatic parameter selection should be taken in order to ensure optimal performance of the eye tracker in any conditions. Furthermore, it was observed that in some cases the eye tracker wrongly detects the inner eye corner as the pupil center. It might be a solution to use the simple template based eye locator in order to reduce the search space for the more accurate isophote based eye tracker. Some simple and naive heuristics regarding the physiological structure of the human face and average distance between the eyes can be incorporated in order to increase the accurate pupil detection;}
\item{head tracking errors - there are errors associated with the head tracker as well. The head tracker accuracy is discussed in \cite{VALENTI&YUCEL&GEVERS}. Since we incorporate the knowledge for the current position of the head in 3D space into our final decision on point of visual gaze it is clear that these errors contribute to the final device error. Furthermore, it was observed that the natural head movement in front of the computer screen when the subject is at $750mm$ distance is not too big. This means that we face the same problem as the one discussed associated with the eye tracker - small errors in the estimate for the current position of the head result in big errors for the final estimation of the point of visual gaze.}
\end{itemize}

The calibration error $\epsilon_{c}$ has several origins as well:
\begin{itemize}
\item{the distance from the screen - the current implementation of the cylindrical head tracker assumes that the system is initialized at $750mm$ from the computer screen. Unfortunately, this is not the case in most scenarios. Since we model the space in front of the user based on this assumption, wrong initialization results in wrong model of that space. One way to cope with this problem is to use visual cues on the computer screen. Since we can already model the space between the user and the screen, we can supply the user with visual cues on the screen regarding his/her current position compared to the model and then the tracking can start once the user's position is as close as possible to the model. Another way is to perform camera calibration step, where the camera intrinsic parameters are estimated and then to model the space in front of the user based on these parameters. Furthermore, currently we use a pin hole camera model for the camera. Though, it has been proven that this model is a good approximation of the underlying camera model, we are faced with the difficulty that every approximation we do in our system contributes a lot to the final error.}
\item{do you really look there? - this questions rises again the discussion on how certain a human can be when saying that he/she looks at a point in pixel dimensions. It has been already argued that due to the physiological structure of the eye, we can not say with 99\% confidence that we are looking at specific pixel on the computer screen; there is an uncertainty window associated with that statement. Further discussion follows when we talk for the foveal error. Just for completeness, it is clear that this introduces errors in the total calibration error;}
\item{model selection - as discussed in the proposed approach, the model selection is an important step in the calibration procedure. Currently we assume two models - a linear and a second order polynomial. Although, we reach good results using these models, it is not clear whether they can capture the behavior of the underlaying function in optimal way. Investigation on new models was a part of this research, though it was mostly on the trial-error basis. A more generic approach to this problem needs to be investigated. We could come up with better mathematical model by conducting series of correctly designed experiments and again use the least squares method to approximate the unknown coefficients of the function. Another approach might be to use more complicated methods for learning the underlying model, say, artificial neural networks to learn the unknown function responsible for the accurate mapping from the vectors in camera coordinate system to points in the screen coordinate system;}
\item{new calibration techniques - it is worth investigating whether the current approach for calibration is sufficient enough. The approach we use for calibration proved to be useful in every system in the literature, though we should not forget that these systems are mainly concerned with estimation of the point of visual gaze when the user's head is static. New calibration techniques useful for dynamic head settings need investigation. A proposition in that direction is to calibrate not just the location of the eyes but the position of the head as well. The experiment might look something like, a point is displayed on the screen and then the user is asked to perform a certain series of moves with his/her head while looking at the point. Then the information of the location of the eyes coupled with the information for the position of the head could be used to reason for the point of visual gaze. This follows our intuition that, while the current approach reasons for the point of visual gaze based just on the information for the location of the eyes (calibration wise) is good for static head settings, it is necessary to introduce new variables into the equation when we consider dynamic head settings, namely, the position of the head (calibration wise). The mentioned experiment can go a step further - we can use a chin-rest and perform the described calibration on steps of say $5^\circ$ - $10^\circ$ of head rotation. Then we can learn a function that can approximate the missing facial feature-pupil center vectors for each degree of rotation for each point on the screen.}
\end{itemize}

There are several insights associated with foveal error $\epsilon_{f}$,
\begin{itemize}
\item{decrease the uncertainty window - during the calibration step we can go even further by displaying a big calibration points on the screen that have a contrasting in color dot (1 pixel) in the middle. We can ask the user to look at the dot while the dimensions of the bigger point gradually decrease with the time elapsed. Once the big point disappears we can assume that the user was looking at the dot under consideration and the recording we have at that point in time for the eyes fixation is as close as possible to the one we are looking for;}
\item{model the uncertainty window - as mentioned at the beginning of the manuscript a third step in the visual gaze estimation pipeline was proposed by \cite{VALENTI&VALENTI}. The author proposes to exploit the fact that people's point of visual gaze tend to be attracted by specific points in the picture under consideration; then, we could model the current picture on the screen as a probability map. For each pixel in the image we associate probability regarding how interesting it is. When the user's point of visual is obtained, the point with highest probability close to the estimated one could be taken as the actual point of visual gaze.}
\end{itemize}

We could start the improvements process by performing an experiment where a real ground truth is collected, reducing the contribution of the calibration error to minimum. Then, we could examine the accuracy of the cylindrical head tracker and the eye tracker and lower the device error. We could examine approaches for modeling the fovea error, which we cannot avoid because it is imposed by the physiological structure of the eye.

\section{Main Conclusions}\label{sec:conclusions}

The main conclusion is built around the usability requirements for the system described at the beginning of the manuscript,

\begin{itemize}
\item{\textbf{be accurate} - unfortunately, we cannot argue that the proposed trackers are precise to minutes of arc. We can report that the system commits a mean error of (56.95, 70.82) pixels in $x$ and $y$ direction, respectively, when the user's head is as static as possible (no chin-rests are used). This corresponds to approximately $1.1^{\circ}$ in $x$ and $1.4^{\circ}$ in $y$ direction when the user is at distance of $750mm$ from the computer screen. Furthermore, we can report that the system has a mean error of (87.18, 103.86) pixels in $x$ and $y$ direction, respectively, under natural head movement, which corresponds to approximately $1.7^{\circ}$ in $x$ and $2.0^{\circ}$ in $y$ direction when the user is at distance of $750mm$ from the computer screen;}
\item{\textbf{be reliable and be robust} - the experiments prove that the proposed system has a constant and repetitive behavior and that it can work under different illumination conditions. The dataset used for testing is highly heterogeneous - there are test subjects with different ethnic backgrounds, different gender, under different illumination conditions, wearing glasses and not. The results show that the system performs with repetitive good accuracy;}
\item{\textbf{be non-intrusive and allow for free head motion} - the proposed system does not cause any harm and discomfort, and that is the main reason for its complexity. Because of the restriction of using any lighting sources, chin-rest, high resolution cameras or any other hardware/prior knowledge that will help in the estimation of the point of visual gaze, it does not require any specific hardware and does not affect the user in any way. The proposed tracker is capable to cope with head movements as well.}
\item{\textbf{not require calibration and have real-time response} - as described, the proposed tracker allows for calibration free system. In the design of the last experiment described in the manuscript, this idea was tested and proved to be working. Just conducting a controlled experiment where accurate facial feature-pupil center vectors are collected and used for calibration will increase the accuracy of the proposed tracker. The system has real-time response and can be used with frame rate of 15 frames per second.}
\end{itemize}

\section{Main Contributions}\label{sec:contributions}

We were faced with the problem of accurate estimation of the human's point of visual gaze when using a personal computer. There were many restrictions imposed on the system in terms of hardware and prior knowledge. Our solution is based in appearance-based computer vision algorithms, that are capable to translate noisy image features from a low resolution image capturing device (web camera) to higher level representation (the location of the eyes and the position of the head). Using this information and a calibration procedure we could infer where the user's point of visual gaze is on the computer screen. Furthermore, we proposed a system that achieves good results under natural head movement, nevertheless the limitations imposed.

We evaluated our solution by conducting three main experiments, where the accuracy of the proposed algorithms was tested in static (no head movement) and dynamic (with head movement) settings. We observed that the proposed tracker improves the accuracy with a factor of about 3. The results prove that the proposed tracker works, regardless the simplicity of the ideas behind it. We found many problems and origins of errors in the estimate, that were discussed in details in section \ref{sec:improvements} where we propose possible solutions for them.

It is obvious that this project is accompanied with tremendous amount of research, and there are important insights drawn while thinking for solution. There are advices and lessons learned conducting this research and it is reasonable to believe that it makes a step towards meeting all the usability requirements for building an ideal visual gaze tracker. Finally, we successfully managed to increase the bandwidth from the user to the computer and to turn the computer from a tool to an agent that is capable to perceive this important sense, enabling more intelligent and intuitive human-computer interaction.
\cleardoublepage

\acknowledgments

\begin{small}
This master thesis has been a challenging and rewarding experience and here I will take the opportunity to express my sincere gratitude to the people that turned this experience into a pleasurable one.

I thank my parents, who always emphasized the importance of good education and who always encouraged me to learn. Next, I thank my brother, who never stops pushing me towards my limits and as such, constantly redefining my personal believes of my capabilities.

I thank my wonderful professor, Theo, for introducing me to the fascinating world of Artificial Intelligence, and for guiding me through it. I thank Roberto, for his invaluable supervision, and for the nice and insightful discussions.

I thank the countless IBM-ers I encountered last several months. The interaction with them got me even more inspired and redefined the meaning of what is possible.

I thank all my friends for always being ready to have a beer when I needed to clear my mind. 

Last but not least, I thank my girlfriend for being the amazing person she is and for being there in the hardest moments.
\\[2ex]
Amsterdam\hfill Kalin Stefanov\\
October 2010
\end{small}
\cleardoublepage


\appendix
\chapter{Least Squares Fitting}\label{chapter:AppendixA}

\begin{figure}[htp!]
\begin{center}
\includegraphics[scale=1]{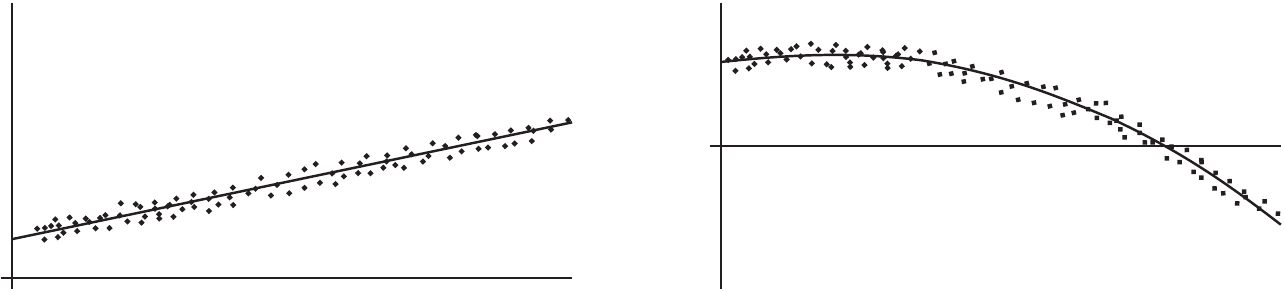}
\caption{Least squares fitting}
\label{fig:leastSquaresFitting1}
\end{center}
\end{figure}

A mathematical procedure for finding the best-fitting curve to a given set of points (Figure \ref{fig:leastSquaresFitting1}) by minimizing the sum of the squares of the offsets (``the residuals'') of the points from the curve \cite{WEISSTEIN1}. The sum of the squares of the offsets is used instead of the offset absolute values because this allows the residuals to be treated as a continuous differentiable quantity. However, because squares of the offsets are used, outlying points can have a disproportionate effect on the fit, a property which may or may not be desirable depending on the problem at hand.

\begin{figure}[htp!]
\begin{center}
\includegraphics[scale=1]{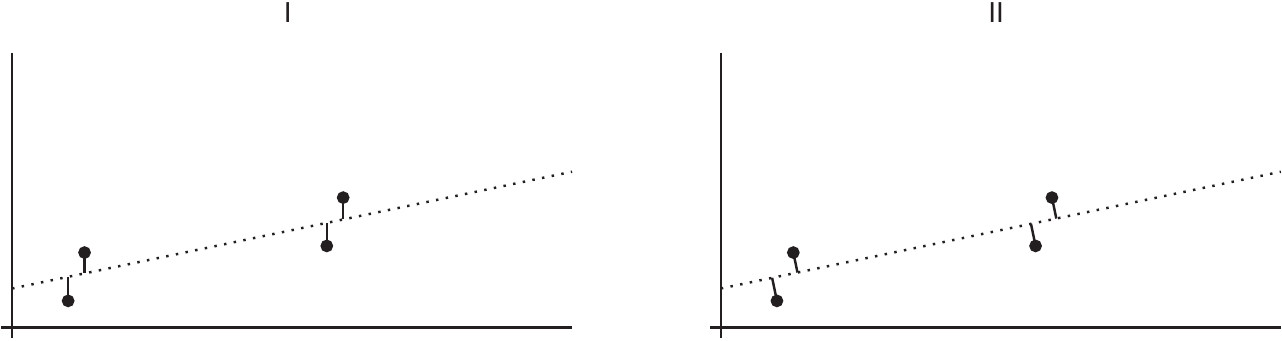}
\caption{Vertical and perpendicular offsets}
\label{fig:leastSquaresFitting2}
\end{center}
\end{figure}

In practice, the vertical offsets (Figure \ref{fig:leastSquaresFitting2} I) from a line (polynomial, surface, hyperplane, etc.) are almost always minimized instead of the perpendicular offsets (Figure \ref{fig:leastSquaresFitting2} II). This provides a fitting function for the independent variable $\mathbf{X}$ that estimates $y$ for a given $x$ (most often what an experimenter wants), allows uncertainties of the data points along the $x$- and $y$-axes to be incorporated simply, and also provides a much simpler analytic form for the fitting parameters than would be obtained using a fit based on perpendicular offsets. In addition, the fitting technique can be easily generalized from a best-fit line to a best-fit polynomial when sums of vertical distances are used. In any case, for a reasonable number of noisy data points, the difference between vertical and perpendicular fits is quite small.

The \textit{linear} least squares fitting technique is the simplest and most commonly applied form of linear regression and provides a solution to the problem of finding the best fitting \textit{straight} line through a set of points. In fact, if the functional relationship between the two quantities being graphed is known to within additive or multiplicative constants, it is common practice to transform the data in such a way that the resulting line \textit{is} a straight line, say by plotting $\mathbf{T}$ vs. $\sqrt{l}$ instead of $\mathbf{T}$ vs. $l$ in the case of analyzing the period $\mathbf{T}$ of a pendulum as a function of its length $l$. For this reason, standard forms for exponential, logarithmic, and power laws are often explicitly computed. The formulas for linear least squares fitting were independently derived by Gauss and Legendre.

For nonlinear least squares fitting to a number of unknown parameters, linear least squares fitting may be applied iteratively to a linearized form of the function until convergence is achieved. However, it is often also possible to linearize a nonlinear function at the outset and still use linear methods for determining fit parameters without resorting to iterative procedures. This approach does commonly violate the implicit assumption that the distribution of errors is normal, but often still gives acceptable results using normal equations, a pseudoinverse, etc. Depending on the type of fit and initial parameters chosen, the nonlinear fit may have good or poor convergence properties. If uncertainties (in the most general case, error ellipses) are given for the points, points can be weighted differently in order to give the high-quality points more weight.

Vertical least squares fitting proceeds by finding the sum of the \textit{squares} of the vertical deviations $R^{2}$ of a set of $n$ data points
\begin{equation}
R^{2} = \displaystyle\sum[y_{i} - f(x_{i},a_{1},a_{2}, ... ,a_{n})]^{2}
\end{equation}
from a function $f$. Note that this procedure does \textit{not} minimize the actual deviations from the line (which would be measured perpendicular to the given function). In addition, although the \textit{unsquared} sum of distances might seem a more appropriate quantity to minimize, use of the absolute value results in discontinuous derivatives which cannot be treated analytically. The square deviations from each point are therefore summed, and the resulting residual is then minimized to find the best fit line. This procedure results in outlying points being given disproportionately large weighting.

The condition for $R^{2}$ to be a minimum is that
\begin{equation}
\frac{\partial(R^{2})}{\partial a_{i}} = 0
\end{equation}
for $i=1, ..., n$. For a linear fit,
\begin{equation}
f(a,b) = a + bx
\end{equation}
so
\begin{equation}
R^{2}(a,b) = \displaystyle\sum_{i=1}^{n}[y_{i} - (a + bx_{i})]^2
\end{equation}
\begin{equation}
\frac{\partial(R^{2})}{\partial a} = -2\displaystyle\sum_{i=1}^{n}[y_{i} - (a + bx_{i})] = 0
\end{equation}
\begin{equation}
\frac{\partial(R^{2})}{\partial b} = -2\displaystyle\sum_{i=1}^{n}[y_{i} - (a + bx_{i})]x_{i} = 0
\end{equation}
These lead to the equations
\begin{equation}
na + b\displaystyle\sum_{i=1}^{n}x_{i} = \displaystyle\sum_{i=1}^{n}y_{i}
\end{equation}
\begin{equation}
a\displaystyle\sum_{i=1}^{n}x_{i} + b\displaystyle\sum_{i=1}^{n}x_{i}^{2} = \displaystyle\sum_{i=1}^{n}x_{i}y_{i}
\end{equation}
In matrix  form,
\begin{equation}
\left[ \begin{array}{cc}
n & \displaystyle\sum_{i=1}^{n}x_{i} \\
\displaystyle\sum_{i=1}^{n}x_{i} & \displaystyle\sum_{i=1}^{n}x_{i}^{2} \end{array} \right] \left[ \begin{array}{c}
a \\
b \end{array} \right] = \left[ \begin{array}{c}
\displaystyle\sum_{i=1}^{n}y_{i} \\
\displaystyle\sum_{i=1}^{n}x_{i}y_{i} \end{array} \right]
\end{equation}
so
\begin{equation}
\left[ \begin{array}{c}
a \\
b \end{array} \right] = \left[ \begin{array}{cc}
n & \displaystyle\sum_{i=1}^{n}x_{i} \\
\displaystyle\sum_{i=1}^{n}x_{i} & \displaystyle\sum_{i=1}^{n}x_{i}^{2} \end{array} \right]^{-1} \left[ \begin{array}{c}
\displaystyle\sum_{i=1}^{n}y_{i} \\
\displaystyle\sum_{i=1}^{n}x_{i}y_{i} \end{array} \right]
\end{equation}
The $2\times2$ matrix inverse is
\begin{equation}
\left[ \begin{array}{c}
a \\
b \end{array} \right] = \frac{1}{n\displaystyle\sum_{i=1}^{n}x_{i}^{2} - \left(\displaystyle\sum_{i=1}^{n}x_{i}\right)^{2}} \left[ \begin{array}{c}
\displaystyle\sum_{i=1}^{n}y_{i}\displaystyle\sum_{i=1}^{n}x_{i}^{2} - \displaystyle\sum_{i=1}^{n}x_{i}\displaystyle\sum_{i=1}^{n}x_{i}y_{i} \\
n\displaystyle\sum_{i=1}^{n}x_{i}y_{i} - \displaystyle\sum_{i=1}^{n}x_{i}\displaystyle\sum_{i=1}^{n}y_{i} \end{array} \right]
\end{equation}
so
\begin{equation}
a = \frac{\displaystyle\sum_{i=1}^{n}y_{i}\displaystyle\sum_{i=1}^{n}x_{i}^{2} - \displaystyle\sum_{i=1}^{n}x_{i}\displaystyle\sum_{i=1}^{n}x_{i}y_{i}}{n\displaystyle\sum_{i=1}^{n}x_{i}^{2} - \left(\displaystyle\sum_{i=1}^{n}x_{i}\right)^2}
\end{equation}
\begin{equation}
= \frac{\overline{y}\left(\displaystyle\sum_{i=1}^{n}x_{i}^{2}\right) - \overline{x}\displaystyle\sum_{i=1}^{n}x_{i}y_{i}}{\displaystyle\sum_{i=1}^{n}x_{i}^{2} - n\overline{x}^{2}}
\end{equation}
\begin{equation}
b = \frac{n\displaystyle\sum_{i=1}^{n}x_{i}y_{i} - \displaystyle\sum_{i=1}^{n}x_{i}\displaystyle\sum_{i=1}^{n}y_{i}}{n\displaystyle\sum_{i=1}^{n}x_{i}^{2} - \left(\displaystyle\sum_{i=1}^{n}x_{i}\right)^{2}}
\end{equation}
\begin{equation}
= \frac{\displaystyle\sum_{i=1}^{n}x_{i}y_{i} - n\overline{x} \overline{y}}{\displaystyle\sum_{i=1}^{n}x_{i}^{2} - n\overline{x}^{2}}
\end{equation}
These can be rewritten in a simpler form by defining the sums of squares
\begin{equation}
ss_{xx} = \displaystyle\sum_{i=1}^{n}(x_{i} - \overline{x})^{2}
\end{equation}
\begin{equation}
= \left(\displaystyle\sum_{i=1}^{n}x_{i}^{2}\right) - n\overline{x}^{2}
\end{equation}
\begin{equation}
ss_{yy} = \displaystyle\sum_{i=1}^{n}(y_{i} - \overline{y})^{2}
\end{equation}
\begin{equation}
= \left(\displaystyle\sum_{i=1}^{n}y_{i}^{2}\right) - n\overline{y}^{2}
\end{equation}
\begin{equation}
ss_{xy} = \displaystyle\sum_{i=1}^{n}(x_{i} - \overline{x})(y_{i} - \overline{y})
\end{equation}
\begin{equation}
= \left(\displaystyle\sum_{i=1}^{n}x_{i}y_{i}\right) - n\overline{x}\overline{y}
\end{equation}
which are also written as
\begin{equation}
\sigma_{x}^{2} = \frac{ss_{xx}}{n}
\end{equation}
\begin{equation}
\sigma_{y}^{2} = \frac{ss_{yy}}{n}
\end{equation}
\begin{equation}
cov(x,y) = \frac{ss_{xy}}{n}
\end{equation}
Here, $cov(x,y)$ is the covariance and $\sigma_{x}^{2}$ and $\sigma_{y}^{2}$ are variances. Note that the quantities $\displaystyle\sum_{i=1}^{n}x_{i}y_{i}$ and $\displaystyle\sum_{i=1}^{n}x_{i}^{2}$  can also be interpreted as the dot products
\begin{equation}
\displaystyle\sum_{i=1}^{n}x_{i}^{2} = \mathbf{x\cdot x}
\end{equation}
\begin{equation}
\displaystyle\sum_{i=1}^{n}x_{i}y_{i} = \mathbf{x\cdot y}
\end{equation}
In terms of the sums of squares, the regression coefficient $b$ is given by
\begin{equation}
b = \frac{cov(x,y)}{\sigma_{x}^{2}} = \frac{ss_{xy}}{ss_{xx}}
\end{equation}
and $a$ is given in terms of $b$ using ($\diamond$) as
\begin{equation}
a = \overline{y} - b\overline{x}
\end{equation}
The overall quality of the fit is then parameterized in terms of a quantity known as the correlation coefficient, defined by
\begin{equation}
r^{2} = \frac{ss_{xy}^{2}}{ss_{xx}ss_{yy}}
\end{equation}
which gives the proportion of $ss_{yy}$ which is accounted for by the regression.
Let $\hat{y}_{i}$ be the vertical coordinate of the best-fit line with $x$-coordinate $x_{i}$, so
\begin{equation}
\hat{y}_{i} \equiv a + bx_{i}
\end{equation}
hen the error between the actual vertical point $y_{i}$ and the fitted point is given by
\begin{equation}
e_{i} \equiv y_{i} - \hat{y}_{i}
\end{equation}
Now define $s^{2}$ as an estimator for the variance in $e_{i}$,
\begin{equation}
s^{2} = \displaystyle\sum_{i=1}^{n}\frac{e^{2}_{i}}{n - 2}
\end{equation}
Then $s$ can be given by
\begin{equation}
s = \sqrt{\frac{ss_{yy} - bss_{xy}}{n-2}} = \sqrt{\frac{ss_{yy}-\frac{ss_{xy}^{2}}{ss_{xx}}}{n-2}}
\end{equation}
The standard errors for $a$ and $b$ are
\begin{equation}
SE(a) = s \sqrt{\frac{1}{n} + \frac{\overline{x}^{2}}{ss_{xx}}}
\end{equation}
\begin{equation}
SE(b) = \frac{s}{\sqrt{ss_{xx}}}
\end{equation}
\cleardoublepage

\chapter{Least Squares Fitting - Polynomial}\label{chapter:AppendixB}

Generalizing from a straight line (i.e., first degree polynomial) to a \textit{k}th degree polynomial \cite{WEISSTEIN2}
\begin{equation}
y = a_{0} + a_{1}x + ... + a_{k}x^{k},
\end{equation}
the residual is given by
\begin{equation}
R^{2} = \displaystyle\sum_{i=1}^{n}[y_{i} - (a_{0} + a_{1}x_{i} + ... + a_{k}x_{i}^{k})]^{2}.
\end{equation} 
The partial derivatives (again dropping superscripts) are
\begin{equation}
\frac{\partial(R^{2})}{\partial a_{0}} = -2\displaystyle\sum_{i=1}^{n}[y - (a_{0} + a_{1}x + ... + a_{k}x^{k})] = 0
\end{equation}
\begin{equation}
\frac{\partial(R^{2})}{\partial a_{1}} = -2\displaystyle\sum_{i=1}^{n}[y - (a_{0} + a_{1}x + ... + a_{k}x^{k})]x = 0
\end{equation}
\begin{equation}
\frac{\partial(R^{2})}{\partial a_{k}} = -2\displaystyle\sum_{i=1}^{n}[y - (a_{0} + a_{1}x + ... + a_{k}x^{k})]x^{k} = 0
\end{equation}
These lead to the equations
\begin{equation}
a_{0}n + a_{1}\displaystyle\sum_{i=1}^{n}x_{i} + ... + a_{k}\displaystyle\sum_{i=1}^{n}x_{i}^{k} = \displaystyle\sum_{i=1}^{n}y_{i}
\end{equation}
\begin{equation}
a_{0}\displaystyle\sum_{i=1}^{n}x_{i} + a_{1}\displaystyle\sum_{i=1}^{n}x_{i}^{2} + ... + a_{k}\displaystyle\sum_{i=1}^{n}x_{i}^{k+1} = \displaystyle\sum_{i=1}^{n}x_{i}y_{i}
\end{equation}
\begin{equation}
a_{0}\displaystyle\sum_{i=1}^{n}x_{i}^{k} + a_{1}\displaystyle\sum_{i=1}^{n}x_{i}^{k+1} + ... + a_{k}\displaystyle\sum_{i=1}^{n}x_{i}^{2k} = \displaystyle\sum_{i=1}^{n}x_{i}^{k}y_{i}
\end{equation}
or, in matrix form
\begin{equation}
\left[ \begin{array}{cccc}
n & \displaystyle\sum_{i=1}^{n}x_{i} & \hdots & \displaystyle\sum_{i=1}^{n}x_{i}^{k} \\
\displaystyle\sum_{i=1}^{n}x_{i} & \displaystyle\sum_{i=1}^{n}x_{i}^{2} & \hdots & \displaystyle\sum_{i=1}^{n}x_{i}^{k+1} \\
\vdots & \vdots & \ddots & \vdots \\
\displaystyle\sum_{i=1}^{n}x_{i}^{k} & \displaystyle\sum_{i=1}^{n}x_{i}^{k+1} & \hdots & \displaystyle\sum_{i=1}^{n}x_{i}^{2k} \end{array} \right] \left[ \begin{array}{c}
a_{0} \\
a_{1} \\
\vdots \\
a_{k} \end{array} \right] = \left[ \begin{array}{c}
\displaystyle\sum_{i=1}^{n}y_{i} \\
\displaystyle\sum_{i=1}^{n}x_{i}y_{i} \\
\vdots \\
\displaystyle\sum_{i=1}^{n}x_{i}^{k}y_{i} \end{array} \right]
\end{equation}
This is a Vandermonde matrix. We can also obtain the matrix for a least squares fit by writing
\begin{equation}
\left[ \begin{array}{cccc}
1 & x_{1} & \hdots & x_{1}^{k} \\
1 & x_{2} & \hdots & x_{2}^{k} \\
\vdots & \vdots & \ddots & \vdots \\
1 & x_{n} & \hdots & x_{n}^{k} \end{array} \right] \left[ \begin{array}{c}
a_{0} \\
a_{1} \\
\vdots \\
a_{k} \end{array} \right] = \left[ \begin{array}{c}
y_{1} \\
y_{2} \\
\vdots \\
y_{n} \end{array} \right]
\end{equation}
Premultiplying both sides by the transpose of the first matrix then gives
\begin{equation}
\left[ \begin{array}{cccc}
1 & 1 & \hdots & 1 \\
x_{1} & x_{2} & \hdots & x_{n} \\
\vdots & \vdots & \ddots & \vdots \\
x_{1}^{k} & x_{2}^{k} & \hdots & x_{n}^{k} \end{array} \right] \left[ \begin{array}{cccc}
1 & x_{1} & \hdots & x_{1}^{k} \\
1 & x_{2} & \hdots & x_{2}^{k} \\
\vdots & \vdots & \ddots & \vdots \\
1 & x_{n} & \hdots & x_{n}^{k} \end{array} \right] \left[ \begin{array}{c}
a_{0} \\
a_{1} \\
\vdots \\
a_{k} \end{array} \right] = \left[ \begin{array}{cccc}
1 & 1 & \hdots & 1 \\
x_{1} & x_{2} & \hdots & x_{n} \\
\vdots & \vdots & \ddots & \vdots \\
x_{1}^{k} & x_{2}^{k} & \hdots & x_{n}^{k} \end{array} \right] \left[ \begin{array}{c}
y_{1} \\
y_{2} \\
\vdots \\
y_{n} \end{array} \right]
\end{equation}
so
\begin{equation}
\left[ \begin{array}{cccc}
n & \displaystyle\sum_{i=1}^{n}x_{i} & \hdots & \displaystyle\sum_{i=1}^{n}x_{i}^{k} \\
\displaystyle\sum_{i=1}^{n}x_{i} & \displaystyle\sum_{x=1}^{n}x_{i}^{2} & \hdots & \displaystyle\sum_{x=1}^{n}x_{i}^{k+1} \\
\vdots & \vdots & \ddots & \vdots \\
\displaystyle\sum_{x=1}^{n}x_{i}^{k} & \displaystyle\sum_{i=1}^{n}x_{i}^{k+1} & \hdots & \displaystyle\sum_{i=1}^{n}x_{i}^{2k} \end{array} \right] \left[ \begin{array}{c}
a_{0} \\
a_{1} \\
\vdots \\
a_{k} \end{array} \right] = \left[ \begin{array}{c}
\displaystyle\sum_{i=1}^{n}y_{i} \\
\displaystyle\sum_{i=1}^{n}x_{i}y_{i} \\
\vdots \\
\displaystyle\sum_{i=1}^{n}x_{i}^{k}y_{i} \end{array} \right]
\end{equation}
As before, given $n$ points ($x_{i}, y_{i}$) and fitting with polynomial coefficients $a_{0}, ..., a_{k}$ gives
\begin{equation}
\left[ \begin{array}{c}
y_{1} \\
y_{2} \\
\vdots \\
y_{n} \end{array} \right] = \left[ \begin{array}{ccccc}
1 & x_{1} & x_{1}^{2} & \hdots & x_{1}^{k} \\
1 & x_{2} & x_{2}^{2} & \hdots & x_{2}^{k} \\
\vdots & \vdots & \vdots & \ddots & \vdots \\
1 & x_{n} & x_{n}^{2} & \hdots & x_{n}^{k} \end{array} \right] \left[ \begin{array}{c}
a_{0} \\
a_{1} \\
\vdots \\
a_{k} \end{array} \right]
\end{equation}
In matrix notation, the equation for a polynomial fit is given by
\begin{equation}
\mathbf{y} = \mathbf{X}\mathbf{a}
\end{equation}
This can be solved by premultiplying by the matrix transpose $\mathbf{X}^\mathsf{T}$,
\begin{equation}
\mathbf{X}^\mathsf{T}\mathbf{y} = \mathbf{X}^\mathsf{T}\mathbf{X}\mathbf{a}
\end{equation}
This matrix equation can be solved numerically, or can be inverted directly if it is well formed, to yield the solution vector
\begin{equation}
\mathbf{a} = (\mathbf{X}^\mathsf{T} \mathbf{X})^{-1}\mathbf{X}^\mathsf{T}\mathbf{y}
\end{equation}
Setting $k=1$ in the above equations reproduces the linear solution.
\cleardoublepage

\chapter{Line-Plane Intersection}\label{chapter:appendixC}

The plane determined by the points $\mathbf{x}_{1}$, $\mathbf{x}_{2}$, and $\mathbf{x}_{3}$ and the line passing through the points $\mathbf{x}_{4}$ and $\mathbf{x}_{5}$ intersect in a point which can be determined by solving the four simultaneous equations:
\begin{equation}
0 = \left[ \begin{array}{cccc}
x & y & z & 1\\
x_{1} & y_{1} & z_{1} & 1 \\
x_{2} & y_{2} & z_{2} & 1 \\
x_{3} & y_{3} & z_{3} & 1 \end{array} \right]
\end{equation}
\begin{equation}
x = x_{4} + (x_{5} - x_{4})t
\end{equation}
\begin{equation}
y = y_{4} + (y_{5} - y_{4})t
\end{equation}
\begin{equation}
z = z_{4} + (z_{5} - z_{4})t
\end{equation}
for $x$, $y$, $z$, and $t$, giving:
\begin{equation}
t = -\frac{\left[ \begin{array}{cccc}
1 & 1 & 1 & 1\\
x_{1} & x_{2} & x_{3} & x_{4} \\
y_{1} & y_{2} & y_{3} & y_{4} \\
z_{1} & z_{2} & z_{3} & z_{4} \end{array} \right]}{\left[ \begin{array}{cccc}
1 & 1 & 1 & 0\\
x_{1} & x_{2} & x_{3} & x_{5} - x_{4}\\
y_{1} & y_{2} & y_{3} & y_{5} - y_{4} \\
z_{1} & z_{2} & z_{3} & z_{5} - z_{4} \end{array} \right]}.
\end{equation}
This value can then be plugged back in to (C.2), (C.3), and (C.4) to give the point of intersection ($x$, $y$, $z$). \cite{WEISSTEIN3}
\cleardoublepage

\clearpage
\nocite{*}
\bibliographystyle{acm}
\bibliography{Bibliography}

\pdfinfo{
   /Title  (Webcam-based Eye Gaze Tracking under Natural Head Movement)
   /Subject (Webcam-based Eye Gaze Tracking under Natural Head Movement)
   /Author (Kalin M. Stefanov)
   /Keywords (AI, HCI, Computer Vision, Tracking, Eyes, Head, Gaze)
}

\end{document}